%% file: main.tex
\documentclass[11pt,a4paper]{article}
\pdfoutput=1
\usepackage[hyperref]{acl2021}
\usepackage{times}
\usepackage{latexsym}

\usepackage{microtype}

\aclfinalcopy 

\usepackage{adjustbox}
\usepackage{booktabs}
\usepackage{multirow}

\usepackage{microtype}

\usepackage{xspace}
\usepackage{mathtools}
\usepackage{amssymb}
\usepackage{footnote}
\usepackage{tablefootnote}
\usepackage{graphicx}
\usepackage{color, colortbl}
\usepackage{xstring}
\usepackage{comment}
\usepackage{adjustbox}
\usepackage{float}
\restylefloat{table}
\usepackage{array,booktabs,makecell}
\usepackage{multirow}
\usepackage{graphicx}
\usepackage{appendix}
\usepackage[hide]{boxnotes}

\usepackage{cancel}
\usepackage{ulem,xpatch}

\xpatchcmd{\sout}
  {\bgroup}
  {\bgroup}
  {}{}

\newcommand{\eat}[1]{}

\usepackage{caption}
\usepackage[font={footnotesize}]{caption}
\newcolumntype{H}{>{\setbox0=\hbox\bgroup}c<{\egroup}@{}}

\usepackage{arydshln}
\def \ourmodel{ARBERT}
\def \newmodel{MARBERT}

\newcommand\blfootnote[1]{%
  \begingroup
  \renewcommand\thefootnote{}\footnote{#1}%
  \addtocounter{footnote}{-1}%
  \endgroup
}

\PassOptionsToPackage{hyphens}{url}\usepackage{hyperref}
    
\title{\ourmodel \ \& \newmodel: Deep Bidirectional Transformers for Arabic}

\author{Muhammad Abdul-Mageed$^{\dagger}$ ~~~ AbdelRahim Elmadany$^{\dagger}$ ~~~ El Moatez Billah Nagoudi$^{\dagger}$ \\
\normalsize Natural Language Processing Lab  \\
  \normalsize The University of British Columbia\\
      
  \texttt{ \small \{muhammad.mageed,a.elmadany,moatez.nagoudi\}@ubc.ca}
  }

\date{}

\begin{document}
\maketitle
\sloppy
\begin{abstract}
  Pre-trained language models (LMs) are currently integral to many natural language processing systems. Although multilingual LMs were also introduced to serve many languages, these have limitations such as being costly at inference time and the size and diversity of non-English data involved in their pre-training. We remedy these issues for a collection of diverse Arabic varieties by introducing two powerful deep bidirectional transformer-based models, ARBERT and MARBERT. To evaluate our models, we also introduce ARLUE, a new benchmark for multi-dialectal Arabic language understanding evaluation. ARLUE is built using $42$ datasets targeting six different task clusters, allowing us to offer a series of standardized experiments under rich conditions. When fine-tuned on ARLUE, our models collectively achieve new state-of-the-art results across the majority of tasks ($37$ out of $48$ classification tasks, on the $42$ datasets). Our best model acquires the highest ARLUE score ($77.40$) across all six task clusters, outperforming all other models including XLM-R\textsubscript{Large} ($\sim 3.4 \times $ larger size). Our models are publicly available at \href{https://github.com/UBC-NLP/marbert}{https://github.com/UBC-NLP/marbert} and ARLUE will be released through the same repository. 
  ~\blfootnote{ $^{\dagger}$ All authors contributed equally.} 
\end{abstract}

\input{intro}
\input{araLMs}
\input{models}
\input{finetuning}
\input{tasks}

\input{ArBench}
\input{lit}

\input{conc}
\input{acknow}
\input{ethics}

\normalem
\bibliography{acl2021}
\bibliographystyle{acl_natbib}

\appendix
\input{appendix.tex}

\end{document}

%% file: intro.tex
\section{Introduction}\label{sec:intro}
Language models (LMs) exploiting self-supervised learning such as BERT~\cite{devlin2019bert} and RoBERTa~\cite{liu2019roberta} have recently emerged as powerful transfer learning tools that help improve a very wide range of natural language processing (NLP) tasks. 
Multilingual LMs such as mBERT~\cite{devlin2019bert} and XLM-RoBERTa (XLM-R)~\cite{conneau-etal-2020-unsupervised} have also been introduced, but are usually outperformed by monolingual models pre-trained with larger vocabulary and bigger language-specific datasets~\cite{virtanen2019multilingual,antoun2020arabert,dadas2020pre,de2019bertje,le2020flaubert,martin-etal-2020-camembert,nguyen2020phobert}. 




Since LMs are costly to pre-train, it is important to keep in mind the end goals they will serve once developed. For example, \textbf{(i)} in addition to their utility on `standard' data, it is useful to endow them with ability to excel on wider real world settings such as in social media. Some existing LMs do not meet this need since they were trained on datasets that do not sufficiently capture the nuances of social media language (e.g., frequent use of abbreviations, emoticons, and hashtags; playful character repetitions; neologisms and informal language). It is also desirable to build models able to \textbf{(ii)} serve diverse communities (e.g., speakers of dialects of a given language), rather than focusing only on mainstream varieties. In addition, once created, models should be \textbf{(iii)} usable in energy efficient scenarios. This means that, for example, medium-to-large models with competitive performance should be preferred to large-to-mega models.

A related issue is \textbf{(iv)} how LMs are evaluated. Progress in NLP hinges on our ability to carry out meaningful comparisons across tasks, on carefully designed benchmarks. Although several benchmarks have been introduced to evaluate LMs, the majority of these are either exclusively in English (e.g.,  DecaNLP~\cite{mccann2018natural}, GLUE~\cite{wang2018glue}, SuperGLUE~\cite{wang2019superglue}) or use machine translation in their training splits (e.g., XTREME~\cite{pmlr-v119-hu20b}). Again, useful as these benchmarks are, this circumvents our ability to measure progress in real-world settings (e.g., training and evaluation on native vs. translated data) for both cross-lingual NLP and in monolingual, non-English environments.  
%

\textbf{Context.} Our objective is to showcase a scenario where we build LMs that meet \textit{all} four needs listed above. That is, we describe novel LMs that (i) excel across domains, including social media, (ii) can serve diverse communities, and (iii) perform well compared to larger (more energy hungry) models (iv) on a novel, standardized benchmark. We choose Arabic as the context for our work since it is a widely spoken language ($\sim 400$M native speakers), with a large number of diverse dialects differing among themselves and from the standard variety, Modern Standard Arabic (MSA). Arabic is also covered by the popular mBERT~\cite{devlin2019bert} and XLM-R~\cite{conneau-etal-2020-unsupervised}, which provides us a setup for meaningful comparisons. That is, not only are we able to empirically measure monolingual vs. multilingual performance under robust conditions using our new benchmark, ARLUE, but we can also demonstrate 
how our base-sized models outperform (or at least are on par with) larger models (i.e., XLM-R\textsubscript{Large}, which is $\sim 3.4 \times$ larger than our models). In the context of our work, we also show how the currently best-performing model dedicated to Arabic, AraBERT~\cite{antoun2020arabert}, suffers from a number of issues. These include (a) not making use of easily accessible data across domains and, more seriously, (b) limited ability to handle Arabic dialects and (c) narrow evaluation. We rectify all these limitations.

\note[mam]{How many individual datasets we now have? Can we prepare a small Excel table with the number of datasets within each task cluster?}

\textbf{Our contributions.} With our stated goals in mind, we introduce \textbf{\ourmodel}~and \textbf{\newmodel}, two Arabic-focused LMs exploiting large-to-massive diverse datasets. For evaluation, we also introduce a novel \textbf{AR}abic
natural \textbf{L}anguage \textbf{U}nderstanding \textbf{E}valuation benchmark (\textbf{ARLUE}). ARLUE is composed of $42$  different datasets, making it by far the largest and most diverse Arabic NLP benchmark we know of. We arrange ARLUE into six coherent cluster tasks and methodically evaluate on each independent dataset as well as each cluster task, ultimately reporting a single ARLUE score. Our models establish new state-of-the-art (SOTA) on the majority of tasks, across all cluster tasks. Our goal is for ARLUE to serve the critical need for measuring progress on Arabic, and facilitate evaluation of multilingual and Arabic LMs. 
To summarize, we offer the following contributions:


\begin{enumerate}
    \item \textbf{We develop ARBERT and MARBERT}, two novel Arabic-specific Transformer LMs pre-trained on very large and diverse datasets to facilitate transfer learning on MSA as well as Arabic dialects.
    \item \textbf{We introduce ARLUE}, a new benchmark developed by collecting and standardizing splits on $42$ datasets across six different Arabic language understanding cluster tasks, thereby facilitating measurement of progress on Arabic and multilingual NLP.
    \item We fine-tune our new powerful models on ARLUE and provide an extensive set of comparisons to available models. \textbf{Our models achieve new SOTA} on all task clusters in $37$ out of $48$ individual datasets and a SOTA \textit{ARLUE score}. 
\end{enumerate}

The rest of the paper is organized as follows:
In Section~\ref{sec:arabicLMs}, we provide an overview of Arabic LMs.  Section~\ref{sec:our_models} describes our Arabic pre-tained models. We evaluate our models on downstream tasks in Section~\ref{sec:tasks}, and present our benchmark ARLUE and evaluation on it in Section~\ref{sec:arbench}. Section~\ref{sec:lit} is an overview of related work.  We conclude in Section~\ref{sec:conclusion}. We now introduce existing Arabic LMs.






%% file: araLMs.tex
\section{Arabic LMs}\label{sec:arabicLMs}

The term \textit{Arabic} refers to a collection of languages, language varieties, and dialects. The standard variety of Arabic is MSA, and there exists a large number of dialects that are usually defined at the level of the region or country~\cite{mageed-etal-2020-nadi,mageed-etal-2021-dialex,mageed-etal-2021-nadi}. A number of Arabic LMs has been developed. The most notable among these is AraBERT~\cite{antoun2020arabert}, which is trained with the same architecture as BERT~\cite{devlin2019bert} and uses the BERT\textsubscript{Base} configuration. 
AraBERT is trained on $23$GB of Arabic text, making $\sim70$M sentences and $3$B words, from Arabic Wikipedia, the Open Source International dataset (OSIAN)~\cite{zeroual2019osian} ($3.5$M news articles from  $24$ Arab countries), and $1.5$B words Corpus from~\newcite{elkhair-2016} ($5$M articles extracted from $10$ news sources).~\newcite{antoun2020arabert} evaluate AraBERT on three Arabic downstream tasks. These are (1) sentiment analysis from six different datasets: HARD~\cite{elnagar2018hotel}, ASTD~\cite{nabil2015astd}, ArsenTD-Lev~\cite{baly2019arsentd}, LABR~\cite{aly2013labr}, and ArSaS~\cite{elmadany2018arsas}. (2) NER, with the ANERcorp~\cite{benajiba2007anersys}, and  (3) Arabic QA, on Arabic-SQuAD and ARCD~\cite{mozannar2019neural} datasets. Another Arabic LM that was also introduced is ArabicBERT~\cite{safaya2020}, which is similarly based on BERT architecture. ArabicBERT was pre-trained on two datasets only, Arabic Wikipedia and Arabic OSACAR~\cite{suarez2019asynchronous}. Since both of these datasets are already included in AraBERT, and Arabic OSACAR\footnote{\href{https://oscar-corpus.com}{https://oscar-corpus.com}.} has significant duplicates, we compare to AraBERT only. GigaBERT~\cite{lan2020empirical}, an Arabic and English LM designed with code-switching data in mind, was also introduced.\footnote{Since GigaBERT is very recent, we could not compare to it. However, we note that our pre-training datasets are much larger (i.e., $15.6$B tokens for \newmodel~vs. $4.3$B Arabic tokens for GigaBERT) and more diverse.}

%% file: models.tex
\section{Our Models}\label{sec:our_models}


\subsection{\ourmodel}\label{subsec:ARBERT}

\subsubsection{Training Data}

We train \ourmodel~ on $61$GB of MSA text ($6.5$B tokens) from the following sources:  

\input{datasets/Train_Data_ARBERT}

\input{tables/ARBERT_data}

\noindent We provide relevant size and token count statistics about the datasets in Table~\ref{tab:ARABERT-DATA}. 

\subsubsection{Training Procedure}


{\bf Pre-processing.} To prepare the raw data for pre-training, we perform light pre-processing. This helps retain a faithful representation of the naturally occurring text. We only remove diacritics and replace URLs, user mentions, and hashtags that may exist in any of the collections with the generic string tokens \texttt{URL}, \texttt{USER}, and \texttt{HASHTAG}, respectively. We do not perform any further pre-processing of the data before splitting the text off to wordPieces~\cite{schuster2012japanese}. 
Multilingual models such as mBERT and XLM-R have $5$K (out of $110$K) and $14$K (out of $250$K) Arabic WordPieces, respectively, in their vocabularies. AraBERT employs a vocabulary of $60$K (out of $64$K).\footnote{The empty $4$K vocabulary bin is reserved for additional wordPieces, if needed.} For \ourmodel, we use a larger vocabulary of $100$K WordPieces. For tokenization, we use the WordPiece tokenizer~\cite{wu2016google} provided by~\newcite{devlin2019bert}. 
 
\noindent {\bf Pre-training.} For \ourmodel,~we follow~\newcite{devlin2019bert}'s pre-training setup. To generate each training input sequence, we use the whole word masking, where $15\%$ of the $N$ input tokens are selected for replacement. These tokens are replaced $80\%$ of the time with the [MASK] token, $10\%$ with a random token, and $10\%$ with the original token. We use the original implementation of BERT in the TensorFlow framework.\footnote{\href{ https://github.com/google-research/bert}{ https://github.com/google-research/bert}.} As mentioned, we use the same network architecture as BERT\textsubscript{Base}: $12$ layers, $768$ hidden units, $12$ heads, for a total of $\sim163$M parameters. We use a batch size of $256$ sequences and a maximum sequence length of $128$ tokens ($256$ sequences $\times$ $128$ tokens = $32,768$ tokens/batch) for $8$M steps, which is approximately $42$ epochs over the $6.5$B tokens. For all our models, we use a learning rate of $1$e$-4$. We pre-train the model on one Google Cloud TPU with eight cores (v$2.8$) from TensorFlow Research Cloud (TFRC).\footnote{\href{https://www.tensorflow.org/tfrc}{https://www.tensorflow.org/tfrc}.} Training took $\sim 16$ days, for $42$ epochs over all the tokens. Table~\ref{tab:models_config_comp} shows a comparison of \ourmodel~ with mBERT, XLM-R, AraBERT, and MARBERT (see Section~\ref{subsec:MARBERT}) in terms of data sources and size, vocabulary size, and model parameters.




\subsection{\newmodel}\label{subsec:MARBERT}

As we pointed out in Sections~\ref{sec:intro} and~\ref{sec:arabicLMs}, Arabic has a large number of diverse dialects. Most of these dialects are under-studied due to rarity of resources. Multilingual models such as mBERT and XLM-R are trained on mostly MSA data, which is also the case for AraBERT and \ourmodel
. As such, these models are not best suited for downstream tasks involving dialectal Arabic. To treat this issue, we use a large Twitter dataset to pre-train a new model, \newmodel, from scratch as we describe next. 

\input{tables/models_config_comp}

\subsubsection{Training data}

To pre-train \newmodel, we randomly sample $1$B Arabic tweets from a large in-house dataset of about $6$B tweets. We only include tweets with at least three Arabic words, based on character string matching, regardless whether the tweet has non-Arabic string or not. That is, we do not remove non-Arabic so long as the tweet meets the three Arabic word criterion. The dataset makes up $128$GB of text ($15.6$B tokens). 

\subsubsection{Training Procedure}

\noindent{\bf Pre-processing.} We employ the same pre-processing as~\ourmodel. 

\noindent {\bf Pre-training.} We use the same network architecture as BERT\textsubscript{Base}, but \textit{without} the next sentence prediction (NSP) objective since tweets are short.\footnote{It was also shown that NSP is \textit{not} crucial for model performance~\cite{liu2019roberta}.} We use the same vocabulary size ($100$K wordPieces) as~\ourmodel, and \newmodel~also has $\sim 160$M parameters. We train \newmodel~ for $17$M steps ($\sim 36$ epochs) with a batch size of $256$ and a maximum sequence length of $128$. Training took $\sim 40$ days on one Google Cloud TPU (eight cores). We now present a comparison between our models and popular multilingual models as well as AraBERT.

\subsection{Model Comparison}
Our models compare to mBERT~\cite{devlin2019bert}, XLM-R~\cite{conneau-etal-2020-unsupervised} (base and large), and AraBERT~\cite{antoun2020arabert} in terms of training data size,  vocabulary size, and overall model capacity as we summarize in Table~\ref{tab:models_config_comp}. In terms of the actual Arabic variety involved,~\newcite{devlin2019bert} train mBERT with Wikipedia Arabic data, which is MSA. XLM-R~\cite{conneau-etal-2020-unsupervised} is trained on Common Crawl data, which likely involves a small amount of Arabic dialects. AraBERT is trained on MSA data only. \ourmodel~ is trained on a large collection of MSA datasets. Unlike all other models, our~\newmodel~model is trained on Twitter data, which involves both MSA and diverse dialects. We now describe our fine-tuning setup.

%% file: datasets/Train_Data_ARBERT.tex
\begin{itemize}

    \item  \textbf{Books (Hindawi)}. We collect and pre-process $1,800$ Arabic books from the public Arabic bookstore Hindawi.\footnote{\href{https://www.hindawi.org/books/}{https://www.hindawi.org/books/}.}

    \item  \textbf{El-Khair}. This is a $5$M news articles dataset from $10$ major news sources covering eight Arab countries from~\newcite{elkhair-2016}.

    \item  \textbf{Gigaword}. We use Arabic Gigaword 5\textsuperscript{th} Edition from the Linguistic Data Consortium (LDC).\footnote{ \href{https://catalog.ldc.upenn.edu/LDC2011T11}{https://catalog.ldc.upenn.edu/LDC2011T11}.} The dataset is a comprehensive archive of newswire text from multiple Arabic news sources.
    \item  \textbf{OSCAR}. This is the MSA and Egyptian Arabic portion of the Open Super-large Crawled Almanach coRpus~\cite{suarez2019asynchronous},\footnote{\href{https://oscar-corpus.com/}{https://oscar-corpus.com/}.} a huge multilingual subset from Common Crawl\footnote{\href{https://commoncrawl.org}{https://commoncrawl.org}.} obtained using language identification and filtering.

    \item  \textbf{OSIAN}. The Open Source International Arabic News Corpus (OSIAN)~\cite{zeroual2019osian} consists of $3.5$ million articles from $31$ news sources in $24$ Arab countries.

    \item  \textbf{Wikipedia Arabic}. We download and use the December $2019$ dump of Arabic Wikipedia. We use WikiExtractor\footnote{\href{https://github.com/attardi/wikiextractor}{https://github.com/attardi/wikiextractor}.} to extract articles and remove markup from the dump.

\end{itemize}

%% file: tables/ARBERT_data.tex
\begin{table}[ht]
\centering
\small
\resizebox{0.85\columnwidth}{!}{%
\begin{tabular}{lcHr}
\hline
\textbf{Source}                          & \textbf{Size}  & \textbf{\#Sentences} & \textbf{\#Tokens} \\ \hline
Books (Hindawi)      & $650$MB                   & $2.3$M            & $72.5$M                                 \\
El-Khair & $16$GB      & $42$M           & $1.6$B                              \\

Gigawords    & $10$GB                    & $67.4$M           & $1.1$B                              \\

OSIAN     & $2.8$GB                   & $12$M           & $292.6$M                                \\

OSCAR-MSA        & $31$GB                    & $67.3$M           & $3.4$B          \\
OSCAR-Egyptian      &$32$MB & $101.9$M              & $3.8$M                                  \\
Wiki     & $1.4$GB                   & $12.5$M           & $156.5$M                                \\\hline
\textbf{Total}                           & \textbf{$\bf 61$GB} & \textbf{$ \bf 203.5$M}                      & \textbf{$\bf 6.5$}B        \\ 

\hline

\end{tabular}%
}
\caption{\ourmodel~'s pre-train resources.}
\label{tab:ARABERT-DATA}
\end{table}

%% file: tables/models_config_comp.tex
\begin{table*}[ht]
\centering
\renewcommand{\arraystretch}{1}
\resizebox{0.8\textwidth}{!}{%
\begin{tabular}{lcrcrHcc}
\toprule
\multicolumn{1}{l}{\multirow{2}{*}{\textbf{Models}}} & \multicolumn{2}{c}{\textbf{Training Data}} & \multicolumn{3}{c}{\textbf{Vocabulary}}& \multicolumn{2}{c}{\textbf{Configuration}} \\  \cline{2-3} \cline{4-7} \cline{7-8}
\multicolumn{1}{c}{}  & \multicolumn{1}{l}{\textbf{Source}}& \multicolumn{1}{c}{\textbf{Tokns} (ar/all)} & \multicolumn{1}{c}{\textbf{Tok}} & \multicolumn{1}{c}{\textbf{Size} (ar/all)} & \textbf{Cased} & \textbf{B / L} & \textbf{Param.}   \\ \toprule

mBERT & Wiki.& $153$M/$1.5$B & WP   & $5$K/$110$K & yes   & B & $110$M  \\ 
XLM-R\textsubscript{B} & CC   & $2.9$B/$295$B  & SP & $14$K/$250$K& yes   & B & $270$M  \\ 
XLM-R\textsubscript{L}& CC   & $2.9$B/$295$B  & SP & $14$K/$250$K & yes   & L& $550$M  \\ 
AraBERT   &  3 sources & $2.5$B/$2.5$B  & SP& $60$K/$64$K & no& B & $135$M  \\  \hdashline 
\textbf{ARBERT}  & 6 sources & $6.2$B/$6.2$B  & WP   & $100$K/$100$K& no& B & $163$M  \\ 
\textbf{MARBERT}& Ara. Tweets  & $15.6$B/$15.6$B& WP   & $100$K/$100$K& no& B & $163$M  \\ \toprule
\end{tabular}%
}

\caption{Models compared. \textbf{B:} Base, \textbf{L:} Large, \textbf{CC:} Common Crawel, \textbf{SP:} SentencePiece, \textbf{WP:} WordPiece.}
\label{tab:models_config_comp} 
\end{table*}

%% file: finetuning.tex
\subsection{Model Fine-Tuning}\label{subsec:finetuning}
We evaluate our models by fine-tuning them on a wide range of tasks, which we thematically organize into six clusters: \textbf{(1)} sentiment analysis (SA), \textbf{(2)} social meaning (SM) (i.e., age and gender, dangerous and hateful speech, emotion, irony, and sarcasm), \textbf{(3)} topic classification (TC), \textbf{(4)} dialect identification (DI), \textbf{(5)} named entity recognition (NER), and \textbf{(6)} question answering (QA). For all classification tasks reported in this paper, we compare our models to four other models: mBERT, XLM-R\textsubscript{Base}, XLM-R\textsubscript{Large}, and AraBERT. We note that XLM-R\textsubscript{Large} is $\sim3.4 \times$ larger than any of our own models ($\sim 550$M parameters vs. $\sim 160$M). We offer two main types of evaluation: on \textbf{\textit{(i) individual tasks}}, which allows us to compare to other works on each individual dataset ($48$ classification tasks on $42$ datasets), and \textit{\textbf{(ii) ARLUE clusters}} (six task clusters).

For all reported experiments, we follow the same light pre-processing we use for pre-training. For all individual tasks and ARLUE task clusters, we fine-tune on the respective training splits for $25$ epochs, identifying the best epoch on development data, and reporting on both development and test data.\footnote{A minority of datasets came with no development split from source, and so we identify and report the best epoch only on test data for these. This allows us to compare all the models under the same conditions ($25$ epochs) and report a fair comparison to the respective original works. For \textit{all} ARLUE cluster tasks, we identify the best epoch \textit{exclusively} on our development sets (shown in Table~\ref{tab:ArBenchdata}).} We typically use the exact data splits provided by original authors of each dataset. Whenever no clear splits are available, or in cases where expensive cross-validation was used in source, we divide the data following a standard $80\%$ training, $10\%$ development, and $10\%$ test split. For all experiments, whether on individual tasks or ARLUE task clusters, we use the Adam optimizer~\cite{kingma2014adam} with input sequence length of $256$, a batch size of $32$, and a learning rate of $2$e$-6$. These values were identified in initial experiments based on development data of a few tasks.\footnote{NER and QA are expetions, where we use sequence lengths of $128$ and $384$, respectively; a batch sizes of $16$ for both; and a learning rate of $2$e$-6$ and $3$e$-5$, respectively.} We now introduce individual tasks.	

%% file: tasks.tex
\section{Individual Downstream Tasks}\label{sec:tasks}

\input{tasks/SA}

\input{tasks/SM}
\input{tasks/TC}
\input{tasks/DIA}
\input{tasks/NER}

\input{tasks/QA}

%% file: tasks/SA.tex
\subsection{Sentiment Analysis}\label{sec:tasks_SA}
\textbf{Datasets.} We fine-tune the language models on all publicly available SA datasets we could find in addition to those we acquired directly from authors. In total, we have the following $17$ MSA and DA datasets: AJGT~\cite{alomari2017arabic}, AraNET\textsubscript{Sent}~\cite{mageed-2020-aranet}, AraSenTi-Tweet~\cite{al2017arasenti}, ArSarcasm\textsubscript{Sent}~\cite{farha2020arabic}, ArSAS~\cite{elmadany2018arsas}, ArSenD-Lev~\cite{baly2019arsentd}, ASTD~\cite{nabil2015astd}, AWATIF~\cite{mageed2012awatif}, BBNS \&  SYTS ~\cite{salameh2015sentiment}, CAMel\textsubscript{Sent}~\cite{obeid2020camel}, HARD~\cite{elnagar2018hotel}, LABR~\cite{aly2013labr}, Twitter\textsubscript{Abdullah}~\cite{abdulla2013arabic},  Twitter\textsubscript{Saad},\footnote{\href{https://www.kaggle.com/mksaad/arabic-sentiment-twitter-corpus}{www.kaggle.com/mksaad/arabic-sentiment-twitter}.} and SemEval-2017~\cite{rosenthal2017semeval}. Details about the datasets and their splits are in Section~\ref{subsection:SA_Datasets}.

\input{tables/tab_SA_res}
\input{tables/tab_SA_acc.tex}
\noindent \textbf{Baselines.} We compare to the STOA listed in Table~\ref{tab:senti_results_I} and Table~\ref{tab:senti_results_II} captions. For all datasets with no baseline in Table~\ref{tab:senti_results_I}, we consider AraBERT our baseline. Details about SA baselines are in Section~\ref{subsection:SA_Baselines}.


\input{evaluation/SA_res}

%% file: tables/tab_SA_res.tex
\begin{table}[ht]
\centering
\renewcommand{\arraystretch}{1.2}
\resizebox{\columnwidth}{!}{%
\begin{tabular}{lccccccc}
\toprule
\textbf{Dataset (classes)} & \textbf{SOTA} &\textbf{mBERT} & \textbf{XLM-R\textsubscript{B}} & \textbf{XLM-R\textsubscript{L}} & \textbf{AraBERT} & \textbf{ARBERT} & \textbf{MARBERT} \\ \toprule
ArSAS (3)       &$92.00$\textsuperscript{$\star$}   & $87.50$ & $90.00$ & $91.50$  & $91.00$   & $92.00$  & $\bf93.00$   \\
ASTD (3)         &$73.00$\textsuperscript{$\star$}     & $67.00$ & $60.67$  & $67.67$  & $72.00$   & $76.50$  & $\bf78.00$  \\
SemEval (3)        &$69.00$\textsuperscript{$\star$}   & $57.00$ & $64.00$  & $67.00$  & $62.00$   & $69.00$  & $\bf71.00$   \\
AraNET\textsubscript{Sent} (2)   &$76.20$\textsuperscript{$\dagger$}     & $84.00$ & $92.00$  & $\bf93.00$  & $86.50$   & $89.00$  & $92.00$   \\
ArSarc\textsubscript{Sent} (3)  & - & $60.50$ & $63.50$  & $70.00$  & $63.50$   & $68.00$  & $\bf71.50$   \\
AraSenTi (3)& -& $89.50$ & $92.00$  & $\bf93.50$  & $91.00$   & $90.00$  & $90.00$   \\
BBN (3)              &- & $55.50$ & $69.50$  & $72.00$  & $70.00$   & $76.50$  & $\bf79.00$   \\
SYTS (3)             &- & $67.00$ & $78.00$  & $76.50$  & $75.50$   & $\bf79.00$  & $76.50$   \\
Tw\textsubscript{Saad} (2)   & -  & $79.00$ & $95.00$  & $95.00$  & $81.00$   & $90.00$  & $\bf96.00$   \\
SAMAR (5)            & -& $22.50$ & $54.00$  & $\bf57.00$  & $36.50$   & $43.50$  & $55.50$   \\
AWATIF (4)           &- & $60.50$ & $63.50$  & $68.50$  & $66.50$   & $71.50$  & $\bf72.50$   \\
Tw\textsubscript{Abdullah} (2) &- & $81.50$ & $91.00$  & $92.00$  & $89.50$   & $91.50$  & $\bf95.00$ \\ 
\toprule

\end{tabular}%
}
\caption{\small{SA results (I) in F\textsubscript{1}\textsuperscript{\textit{PN}}. \textsuperscript{$\star$}~\newcite{obeid2020camel}; \textsuperscript{$\dagger$}~\newcite{mageed-2020-aranet}. Default baseline is AraBERT.}}
\label{tab:senti_results_I}

\end{table}

%% file: tables/tab_SA_acc.tex
\begin{table}[]
\centering
\resizebox{\columnwidth}{!}{%
\renewcommand{\arraystretch}{1.2}
\begin{tabular}{lccccccc}
\toprule
\textbf{Dataset (classes)} & \textbf{SOTA} &\textbf{mBERT} & \textbf{XLM-R\textsubscript{B}} & \textbf{XLM-R\textsubscript{L}} & \textbf{AraBERT} & \textbf{ARBERT} & \textbf{MARBERT} \\ \toprule
AJGT (2)                        &   $93.80$                & $86.67$                               & $89.44$                                & $91.94$                                & $92.22$                                 & $94.44$                                & $\bf96.11$                        \\ 
HARD (2)                        &            $96.20$       & $95.54$                               & $95.74$                                & $95.96$                                & $95.89$                                 & $96.12$                                & $\bf96.17$                        \\ 
ArsenTD-LEV (5)                &      $59.40$              & $50.50$                                & $55.25$                                & $\bf62.00$                          & $56.13$                                 & $61.38$                                & $60.38$                                 \\ 
LABR (2)                 &  $86.70$                       & $91.20$                                & $91.23$                                & $92.20$                                 & $91.97$                                 & $\bf92.51$                                & $92.49$                        \\ 
ASTD-B(2)                   &$92.60$                       & $79.32$                               & $87.59$                                & $77.44$                                & $83.08$                                 & $93.23$                                & $\bf96.24$                        \\ \toprule

\end{tabular}%
}

\caption{\small{SA results (II) in Acc. SOTA by \newcite{antoun2020arabert}.}}
\label{tab:senti_results_II}
\end{table}

%% file: evaluation/SA_res.tex

\noindent\textbf{Results.} To facilitate comparison to previous works with the appropriate evaluation metrics, we split our results into two tables: We show results in F\textsubscript{1}\textsuperscript{PN} in Table~\ref{tab:senti_results_I} and F\textsubscript{1} in Table~\ref{tab:senti_results_II}. We typically \textbf{bold} the best result on each dataset. \textit{\textbf{Our models achieve best results in $13$ out of the $17$ classification tasks reported in the two tables combined}}, while XLM-R (which is a much larger model) outperforms our models in the $4$ remaining tasks. We also note that XLM-R acquires better results than AraBERT in the majority of tasks, a trend that continues for the rest of tasks. Results also clearly show that~\newmodel~is more powerful than than~\ourmodel. This is due to~\newmodel's larger and more diverse pre-training data, especially that many of the SA datasets involve dialects and come from social media.

 %

%% file: tasks/SM.tex

\subsection{Social Meaning Tasks} \label{sec:tasks_SM}

We collectively refer to a host of tasks as \textbf{social meaning}. These are age and gender detection; dangerous, hateful, and offensive speech detection; emotion detection; irony detection; and sarcasm detection. We now describe datasets we use for each of these tasks.


\noindent\textbf{Datasets.} For both age and gender, we use Arap-Tweet~\cite{zaghouani2018arap}. We use AraDan~\cite{alshehri2020understanding} for dangerous speech. For offensive language and hate speech, we use the dataset released in the shared task (sub-tasks A and B)  of offensive speech by~\newcite{mubarak-etal-2020-overview}. We also use AraNET\textsubscript{Emo}~\cite{mageed-2020-aranet}, IDAT@FIRE2019~\cite{idat2019}, and ArSarcasm~\cite{farha2020arabic} for emotion, irony and sarcasm, respectively. More information about these datasets and their splits is  in Appendix~\ref{subsection:SM_Datasets}.



\noindent\textbf{Baselines.} Baselines for social meaning tasks are the SOTA listed in Table~\ref{tab:SM_res} caption. Details about each baseline is in Appendix~\ref{subsection:SM_Baselines}.


%

\input{evaluation/SM_res}

\input{tables/tab_SM_res}

%% file: evaluation/SM_res.tex
\noindent\textbf{Results.} As Table~\ref{tab:SM_res} shows, our models acquire best results on all eight tasks. Of these, \newmodel~achieves best performance on seven tasks, while \ourmodel~is marginally better than \newmodel~on one task  (irony@FIRE2019). \textit{\textbf{The sizeable gains \newmodel~ achieves reflects its superiority on social media tasks. 
 On average, our models are $\bf 9.83$ F$_1$ better than all previous SOTA.}}

%% file: tables/tab_SM_res.tex
\begin{table}[]
\centering


\resizebox{\columnwidth}{!}{%
\renewcommand{\arraystretch}{1.2}
\begin{tabular}{lHccccccc} 

\toprule
\textbf{Task (classes)} & & \textbf{SOTA }& \textbf{mBERT} & \textbf{\textbf{XLM-R\textsubscript{B}}} & \textbf{\textbf{XLM-R\textsubscript{L}}} & \textbf{AraBERT} & \textbf{ARBERT} & \textbf{MARBERT} \\\toprule
Age (3)        & Arab\_Tweet   &   $51.42$ $\ddagger\ddagger$   & $56.35$  & $59.73$   &  $53.60$   & $57.72$    & $58.95$   & $\bf{62.27}$   \\
Dangerous (2)  & Dangerous      &    $59.60$ $\dagger$ & $62.66$  & $62.76$   & $65.01$   & $64.37$    & 63.21   & \textbf{67.53}   \\
Emotion (8)    & AraNETEmo    &    $60.32$ $\ddagger\ddagger$   & $65.79$  & $70.67$   & $74.89$   & $65.68$    & $67.73$   & \textbf{75.83}   \\
Gender (2)     & Arab\_Tweet   & $65.30$ $\ddagger\ddagger$ & $68.06$  & $71.00$   & $71.14$   & $67.75$    & $69.86$   & \textbf{$\bf72.62$}   \\
Hate (2) & OSACT-B       &   $82.28^{**}$  & $72.81$  & $71.33$   & $79.31$   & $78.89$    & $83.02$   & \textbf{84.79}   \\

Irony (2)      & FIRE2019      & $82.40$ $\ddagger$ & $80.96$  & $81.97$   & $82.52$ & $83.01$    & $\bf 85.59$   & $85.33$   \\
Offensive (2)  & OSACT-A       &  $90.51^*$ & $84.25$  & $85.26$   & $88.28$   & $86.57$    & $90.38$   & \textbf{92.41}   \\
Sarcasm (2)    & AraSarcasm   &     $46.60$ $\ddagger\ddagger$  & $68.20$  & $66.76$   & $69.23$   &$ 72.23$    & $75.04$   & \textbf{76.30}   \\

\toprule


\\      

\end{tabular}%
}
\caption{\footnotesize {Results on social meaning tasks. $F_1$ score is the evaluation metric.
\textsuperscript{$\star$}~\newcite{hassan-etal-2020-alt}, 
\textsuperscript{$\star\star$}~\newcite{djandji-etal-2020-multi}, 
\textsuperscript{$\ddagger$} ~~\newcite{zhang2019multi}, \textsuperscript{$\dagger$} ~\newcite{alshehri2020understanding},  \textsuperscript{$\dagger\dagger$}~\newcite{farha2020arabic},  \textsuperscript{$\ddagger\ddagger$}~\newcite{mageed-2020-aranet}}.}
\label{tab:SM_res}

\end{table}

%% file: tasks/TC.tex
\subsection{Topic Classification} \label{sec:tasks_TC}
Classifying documents by topic is a classical task that still has practical utility. We use four TC datasets, as follows:

\noindent\textbf{Datasets.} We fine-tune on Arabic News Text (ANT)~\cite{chouigui2017ant} under three pre-taining settings (\textit{title only}, \textit{text only}, and \textit{title+text}.), Khaleej~\cite{abbas2011evaluation}, and OSAC~\cite{saad2010osac}. Details about these datasets and the classes therein are in Appendix~\ref{subsection:TC_Datasets}.



\noindent\textbf{Baselines.} Since, to the best of our knowledge, there are no published results exploiting deep learning on TC, we consider AraBERT a strong baseline.
\input{evaluation/TC_res}

\input{tables/tab_TC_res}

%% file: evaluation/TC_res.tex

\noindent\textbf{Results.} As Table~\ref{tab:TC_res} shows, \textit{\textbf{\ourmodel~acquires best results on both OSAC and Khaleej, and the title-only setting of ANT.}} AraBERT slightly outperforms our models on the text-only and title+text settings of ANT. 
 


%% file: tables/tab_TC_res.tex
\begin{table}[]

\footnotesize
\resizebox{\columnwidth}{!}{%
\renewcommand{\arraystretch}{1.2}
\begin{tabular}{lcccccc}
\toprule
\multicolumn{1}{c}{\textbf{Dataset (classes)}}  & \textbf{mBERT} & \textbf{\textbf{XLM-R\textsubscript{B}}} & \textbf{\textbf{XLM-R\textsubscript{L}}} & \textbf{AraBERT} & \textbf{ARBERT} & \textbf{MARBERT} \\
\toprule

ANTText (5) &  $        84.89  $ &$ 85.77   $ &$ 86.72   $ &$ \textbf{88.17}  $ &$ 86.87   $ &$ 85.27     $ \\
ANTTitle (5)      &  $78.29  $ &$ 79.96   $ &$ 81.25   $ &$ 81.03    $ &  $\bf81.70$ &  $ 81.19     $ \\
ANTText+Title (5)  &  $  84.67  $ & $ 86.21   $ &$ 86.96   $ &$\bf87.22$ & $ 87.21   $ &  $ 85.60   $ \\
Khallej (4) &$        92.81  $ &$ 91.87   $ &$ 93.56   $ &$ 93.83    $ &$ \textbf{94.53} $ &$ 93.63     $ \\
OSAC (10)   &$        96.84  $ &$ 97.15   $ &$ \bf98.20   $ &$ 97.03    $ &$ 97.50 $ &$ 97.23     $ \\

\toprule
\end{tabular}
}
\caption{\footnotesize{Performance on TC tasks. Our baseline is AraBERT.}}\label{tab:TC_res}
\end{table}

%% file: tasks/DIA.tex
\subsection{Dialect Identification}\label{sec:tasks_DIA}


Arabic dialect identification can be performed at different levels of granularity, including binary (i.e., MSA-DA), regional (e.g., \textit{Gulf}, \textit{Levantine}), country level (e.g., \textit{Algeria}, \textit{Morocco}), and recently province level (e.g., the Egyptian province of \textit{Cairo}, the Saudi province of \textit{Al-Madinah})~\cite{mageed-etal-2020-nadi,mageed-etal-2021-nadi}. 

\noindent\textbf{Datasets.}~~We fine-tune our models on the following datasets: Arabic Online Commentary (AOC)~\cite{zaidan2014arabic}, ArSarcasm\textsubscript{Dia}~\cite{farha2020arabic},\footnote{ArSarcasm\textsubscript{Dia} carries \textit{regional} dialect labels.} MADAR (sub-task 2)~\cite{bouamor2019madar}, NADI-2020~\cite{mageed-etal-2020-nadi},  and QADI~\cite{abdelali2020arabic}. Details about these datasets are in Table~\ref{tab:DA_distrib}.




\noindent\textbf{Baselines.} Our baselines are marked in Table~\ref{tab:DIA_results} caption. Details about the baselines are in Table~\ref{tab:appedix_DIA_res}. 


\input{evaluation/DIA_res}

%% file: evaluation/DIA_res.tex

\noindent\textbf{Results.} 
As Table~\ref{tab:DIA_results} shows, our models outperform all SOTA as well as the baseline AraBERT across all classification levels with sizeable margins. \textit{\textbf{These results reflect the powerful and diverse dialectal representation of \newmodel, enabling it to serve wider communities}}. 
Although \ourmodel~is developed mainly for MSA, it also outperforms all other models. 

\input{tables/tab_DIA_res}

%% file: tables/tab_DIA_res.tex
\begin{table}[]
\centering
 \renewcommand{\arraystretch}{1.2}
\resizebox{\columnwidth}{!}{%
\begin{tabular}{llccccccc}
\toprule
\textbf{Dataset (classes)} & \textbf{Task} & \textbf{SOTA} &\textbf{mBERT} & \textbf{XLM-R\textsubscript{B}} & \textbf{XLM-R\textsubscript{L}} & \textbf{AraBERT} & \textbf{ARBERT} & \textbf{MARBERT} \\ \toprule
ArSarc\textsubscript{Dia}(5) &Regoin    &-   & $43.81$ & $41.71$  & $41.83$  & $47.54$   & $\bf54.70$  &  $51.27$   \\
MADAR (21)     &Country  & - & $34.92$ & $35.91$  & $35.14$  & $34.87$   & $37.90$   & $\bf40.40$    \\
AOC (4)           &Region   &$82.45$\textbf{\textsuperscript{$\star$}}    & $77.27$ & $77.34$  & $78.77$  & $79.20$    & $81.09$  & $\bf82.37$   \\
AOC (3)           & Region  &$78.81$\textbf{\textsuperscript{$\star$}}     &$85.76$ & $86.39$  & $87.56$  & $87.68$   & $89.06$  & $\bf90.85$   \\
AOC (2)           &Binary   &$87.23$\textbf{\textsuperscript{$\star$}}    &  $86.19$ & $86.85$  & $87.30$   & $87.76$   & $88.46$  & $\bf88.59$   \\
QADI (18)         &Country  &$60.60$\textbf{\textsuperscript{$\dagger$}}   & $66.57$ & $77.00$     & $82.73$  & $72.23$   & $88.63$  & $\bf90.89$   \\
NADI (21)         &Country  &$26.78$\textbf{\textsuperscript{$\ddagger$}}   & $13.32$ & $16.36$  & $17.17$  & $17.46$   & $22.56$  & $\bf29.14$   \\
NADI (100)        &Province &$06.06$\textbf{\textsuperscript{$\dagger\dagger$}}  &  $02.13$  & $04.12$   & $5.30$   & $03.13$    & $06.10$    & $\bf06.28$  \\
\toprule

\end{tabular}%
}
\caption{\small{DIA results in F$_1$. \textsuperscript{$\star$}~\newcite{elaraby-abdul-mageed-2018-deep},
\textbf{\textsuperscript{$\dagger$}}~\newcite{abdelali2020arabic}, \textsuperscript{$\ddagger$}~\newcite{mekki:2020:weighted}, \textbf{\textsuperscript{$\dagger\dagger$}} \newcite{talafha2020multi}. Default baseline is AraBERT. }}
\label{tab:DIA_results}
\end{table}

%% file: tasks/NER.tex
\subsection{Named Entity Recognition}\label{sec:tasks_NER}
We fine-tune the models on five NER datasets.


\noindent\textbf{Datasets.} We use ACE03NW and ACE03BN \cite{mitchell2004tides}, ACE04NW~\cite{mitchell2004tides}, ANERcorp~\cite{benajiba2007anersys}, and TW-NER~\cite{darwish2013named}. Table~\ref{tab:NER_distrib} shows the distribution of named entity classes across the five datasets.

\noindent\textbf{Baseline.} We compare our results with SOTA presented by~\newcite{khalifa2019character} and follow them in focusing on person (PER), location (LOC) and organization (ORG) named entity labels while setting other labels to the unnamed entity (O). Details about~\newcite{khalifa2019character} SOTA models are in Appendix~\ref{subsec_app:NER_baseline}.


\input{tables/tab_NER_res}

\input{evaluation/NER_res}

%% file: tables/tab_NER_res.tex
\begin{table}[]
\centering
 \resizebox{\columnwidth}{!}{%
 \renewcommand{\arraystretch}{1.2}
\begin{tabular}{lccccccc}
\toprule
\multicolumn{1}{l}{\textbf{Dataset}} & \textbf{SOTA}  & \textbf{mBERT} & \textbf{XLM-R\textsubscript{B}} & \textbf{XLM-R\textsubscript{L}} & \textbf{AraBERT} & \textbf{ARBERT} & \textbf{MARBERT} \\
\toprule 

ANERcorp   & $ 88.77 $ & $ 86.78 $ & $ 87.24  $ & $ \textbf{89.94} $ & $ 89.13$    & $ 84.38  $ & $ 80.64   $ \\
ACE04NW  & $ \textbf{91.47}$ & $ 86.37 $ & $ 89.93  $ & $ 89.89  $ & $ 89.03$    & $ 88.24  $ & $ 85.02    $ \\
ACE03BN   & $ 94.92 $ & $ 91.23 $ & $ 53.97  $ & $ 85.41  $ & $ 91.94$   & $ \textbf{96.18} $ & $ 79.05    $ \\
ACE03NW   & $ \textbf{91.20}$ & $ 81.40 $ & $ 87.24  $ & $ 90.62  $ & $ 88.09$    & $ 90.09  $ & $ 87.76    $ \\
TW-NER     & $ 65.34 $ & $ 36.83 $ & $ 49.16  $ & $ 54.44  $ & $ 41.26   $ & $ 59.17  $ & $ \textbf{66.67}  $ \\
\toprule


\end{tabular}

 }

\caption{\footnotesize{NER results in F$_1$. SOTA by~\newcite{khalifa2019character}.}}\label{tab:NER_res_1}
\end{table}

%% file: evaluation/NER_res.tex
\noindent\textbf{Results.} As Table~\ref{tab:NER_res_1} shows, our models outperform SOTA on two out of the five NER datasets. We note that even though  SOTA~\cite{khalifa2019character} employ a complex combination of CNNs and character-level LSTMs, which may explain their better results on two datasets,\textit{\textbf{~\newmodel~still achieves highest performance on the social media dataset (TW-NER)}}.


%% file: tasks/QA.tex

\subsection{Question Answering}\label{subsec:tasks_QA}

\noindent\textbf{Datasets.} We use ARCD \cite{mozannar2019neural} and the three \textit{human} translated Arabic test sections of the XTREME benchmark~\cite{pmlr-v119-hu20b}: MLQA~\cite{lewis2019mlqa}, XQuAD~\cite{artetxe2020cross}, and TyDi QA~\cite{artetxe2020cross}. Details about these datasets are in Table~\ref{tab:QA_distrib}.



\noindent\textbf{Baselines.} We compare to ~\newcite{antoun2020arabert} and consider their system a baseline on ARCD. We follow the same splits they used where we fine-tune on Arabic SQuAD~\ \cite{mozannar2019neural} \textit{and}  $50\%$ of ARCD and test on the remaining $50\%$ of ARCD (ARCD-test). For  all other experiments, we fine-tune on the Arabic \textit{machine translated} SQuAD (AR-XTREME) from the \textit{XTREME} multilingual benchmark ~\cite{pmlr-v119-hu20b} and test on the \textit{human translated} test sets listed above. Our baselines in these is ~\newcite{pmlr-v119-hu20b}'s mBERT\textsubscript{Base} model on \textit{gold} (human) data.

\input{tables/tab_QA_res}

\input{evaluation/QA_res}

%% file: tables/tab_QA_res.tex
\begin{table*}[]
\centering
\resizebox{\textwidth}{!}{%
\begin{tabular}{lcccccccccccccccc}
\toprule
\multirow{2}{*}{\textbf{Dataset}} & \multicolumn{2}{c}{\textbf{SOTA}} & \multicolumn{2}{c}{\textbf{mBERT}} & \multicolumn{2}{c}{\textbf{XLM-R\textsubscript{B}}} & \multicolumn{2}{c}{\textbf{XLM-R\textsubscript{L}}} & \multicolumn{2}{c}{\textbf{AraBERT}} & \multicolumn{2}{c}{\textbf{ARBERT}} & \multicolumn{2}{c}{\textbf{MARBERT}} & \multicolumn{2}{c}{\textbf{MARBERT(v2)}} \\ \cline{2-17} 
    & \textbf{EM}  & $\bf F_1$  & \textbf{EM}   & $\bf F_1$  & \textbf{EM}   & $\bf F_1$   & \textbf{EM}   & $\bf F_1$   & \textbf{EM}    & $\bf F_1$   & \textbf{EM}   & $\bf F_1$   & \textbf{EM}    & $\bf F_1$   & \textbf{EM} & $\bf F_1$    \\ \toprule
    
ARCD-test\textsuperscript{$\star$}  & $30.10$\textsuperscript{$\dagger$}  & $61.20$\textsuperscript{$\dagger$}   & $29.63$& $60.93$       & $30.20$&	$59.55$	&$32.05$&	$64.77$& $30.20$ & $62.30$& $30.34$& $63.89$& $21.65$ & $54.06$& $\bf36.75$   & $\bf68.86$  \\ \hline
ARCD-test   & -  & -   & $26.64$& $58.86$       & $27.31$	&$59.61$&	$28.11$&$62.08$& $25.64$ & $59.92$& $27.21$& $60.73$& $23.22$ & $55.14$& $\bf29.63$   & $\bf63.05 $ \\
AR-MLQA     &  $39.00 $\textsuperscript{$\ddagger$}   & $58.90$ \textsuperscript{$\ddagger$}    & $32.93$& $51.57$       & $32.93$& $53.35$& $38.11$& $\bf60.00$& $35.43$ & $55.42$& $34.15$& $53.65$& $28.02$ & $45.14$& $\bf39.23$   & $59.39$  \\   
AR-XQuAD    &$54.20 $\textsuperscript{$\ddagger$}   & $71.00$ \textsuperscript{$\ddagger$}   & $48.66$& $66.26$       & $45.88$& $64.91$&$51.85$& $72.19$& $51.60$ & $68.79$& $49.92$& $67.90$& $41.09$ & $58.46$& $\bf56.55$   & $\bf72.48$  \\ 
AR-TyiDQA   &  $39.00$\textsuperscript{$\ddagger$}   & $58.90$\textsuperscript{$\ddagger$}   & $46.36$& $64.02$       & $39.41$& $60.99$& $44.41$& $67.06$& $44.19$ & $64.39$& $46.80$& $66.94$& $38.98$ & $57.51$& $\bf47.45$   & $\bf67.67$  \\ 
\toprule
\end{tabular}%
}
\caption{QA results. \textsuperscript{$\star$} Results on this test set are with models using the same training data as~\newcite{antoun2020arabert}, while rest of rows report models trained with AR-XTREME~\cite{pmlr-v119-hu20b}. \textsuperscript{$\dagger$} \newcite{antoun2020arabert}; \textsuperscript{$\ddagger$}~\newcite{pmlr-v119-hu20b}.}
\label{tab:QA_res}
\end{table*}

%% file: evaluation/QA_res.tex
\noindent\textbf{Results.} As is standard, we report QA results in terms of both Exact Match (EM) and F$_1$. We find that results with \ourmodel~and~\newmodel~on QA are not competitive, a clear discrepancy from what we have observed thus far on other tasks. We hypothesize this is because the two models are pre-trained with a sequence length of only $128$, which does not allow them to sufficiently capture both a question and its likely answer within the same sequence window during the pre-training.\footnote{In addition,~\newmodel~is not trained on Wikipedia data from where some questions come.} To rectify this, we further pre-train the stronger model,~\newmodel, on the same MSA data as~\ourmodel~in addition to AraNews dataset~\cite{nagoudi2020machine} ($8.6$GB), but with a bigger sequence length of $512$ tokens for $40$ epochs. We call this further pre-trained model \textbf{MARBERT-v2}, noting it has $29$B tokens. As Table~\ref{tab:QA_res} shows, \textit{\textbf{MARBERT-v2 acquires best performance on all but one test set}}, where XLM-R\textsubscript{Large} marginally outperforms us (only in F$_1$).

%% file: ArBench.tex
\section{ARLUE}\label{sec:arbench}
\subsection{ARLUE Categories}\label{subsec:arbench_cats}
We concatenate the corresponding splits of the individual datasets to form \textit{\textbf{ARLUE}}, which is a conglomerate of task clusters. That is, we concatenate all training data from each group of tasks into a single TRAIN, all development into a single DEV, and all test into a single TEST. One exception is the social meaning tasks whose data we keep independent (see \textbf{ARLUE\textsubscript{SM}} below). Table~\ref{tab:ArBenchdata} shows a summary of the ARLUE datasets.\footnote{Again, ARLUE\textsubscript{SM} datasets are kept independent, but to provide a summary of all ARLUE datasets we collate the numbers in Table~\ref{tab:ArBenchdata}.} We now briefly describe how we merge individual datasets into ARLUE.

\begin{table}[h!]
\centering
\resizebox{\columnwidth}{!}{%
\begin{tabular}{lccrrr}
\hline
\textbf{Dataset} & \textbf{\#Datasets}& \textbf{Task} &\textbf{TRAIN}   & \textbf{DEV} & \textbf{TEST}             \\ \hline 
ARLUE\textsubscript{Senti} & $17$& SA & $190.9$K & $6.5$K  & $44.2$K  \\  

ARLUE\textsubscript{SM}\textsuperscript{\textit{$\star$}} &  $8$&SM & $1.51$M  &  $162.5$K  & $166.1$K  \\
ARLUE\textsubscript{Topic} &  $5$&TC & $47.5K$ & $5.9$K  & $5.9$K  \\  
ARLUE\textsubscript{Dia-B} &  $2$&DI & $94.9$K & $10.8$K  & $12.9$K  \\  
ARLUE\textsubscript{Dia-R} &   $2$& DI& $38.5$K & $4.5$K  & $5.3$K   \\  
ARLUE\textsubscript{Dia-C} &  $3$&DI & $711.9$K & $31.5$K  & $52.1$K  \\
ARLUE\textsubscript{NER\textsuperscript{$\dagger$}}&  $5$& NER & $227.7$K& $44.1$K & $66.5$K  \\

ARLUE\textsubscript{QA}\textsuperscript{\textit{$\ddagger$}} & $4$& QA & $101.6$K  &  $517$  & $7.45$K  \\

\toprule
\end{tabular}%
}
\caption{\small{ARLUE categories across the different data splits.  $^{\star}$ Refer to Table~\ref{tab:SM_distrib} for details about individual social meaning datasets in ARLUE\textsubscript{SM}. $^{\dagger}$~Statistics are at the token level. $^{\ddagger}$~Number of question-answer pairs. }  }
\label{tab:ArBenchdata}
\end{table}
\noindent \textbf{ARLUE\textsubscript{Senti}.} To construct ARLUE\textsubscript{Senti}, we collapse the labels \textit{very negative} into \textit{negative}, \textit{very positive} into \textit{positive}, and \textit{objective} into \textit{neutral}, and remove the \textit{mixed} class. This gives us the $3$ classes \textit{negative}, \textit{positive}, and \textit{neutral} for ARLUE\textsubscript{Senti}. Details are in Table ~\ref{tab:SA_distrib}.

\noindent\textbf{ARLUE\textsubscript{SM}.} We refer to the different social meaning datasets collectively as ARLUE\textsubscript{SM}. We do not merge these datasets to preserve the conceptual coherence specific to each of the tasks. Details about individual datasets in ARLUE\textsubscript{SM} are in~\ref{tab:SM_distrib}. 


\noindent\textbf{ARLUE\textsubscript{Topic.}} We straightforwardly merge the TC datasets to form ARLUE\textsubscript{Topic}, without modifying any class labels. Details of ARLUE\textsubscript{Topic} data are in Table~\ref{tab:TC_distrib}. 

\noindent\textbf{ARLUE\textsubscript{Dia.}} We construct three ARLUE\textsubscript{Dia} categories. Namely, we concatenate the AOC and AraSarcasm\textsubscript{Dia} MSA-DA classes to form \textit{\textbf{ARLUE\textsubscript{Dia-B}}} (binary) and the region level classes from the same two datasets to acquire \textit{\textbf{ARLUE\textsubscript{Dia-R}}} (4-classes, \textit{region}). We then merge the country classes from the rest of datasets to get \textit{\textbf{ARLUE\textsubscript{Dia-C}}} (21-classes, \textit{country}). Details are in Table \ref{tab:DA_distrib}. \\
\noindent\textbf{ARLUE\textsubscript{NER} \&}  \textbf{ARLUE\textsubscript{QA}}.  We straightforwardly concatenate all corresponding splits from the different NER and QA datasets to form \textbf{\textit{ARLUE\textsubscript{NER}}} and \textit{\textbf{ARLUE\textsubscript{QA}}}, respectively. Details of each of these task clusters data are in Tables \ref{tab:NER_distrib} and \ref{tab:QA_distrib}, respectively.  

\subsection{Evaluation on ARLUE}\label{subsec:AraLU_eval}
We present results on each task cluster independently using the relevant metric for both the development split (Table~\ref{tab:appendix_Arbench_res}) and test split (Table~\ref{tab:Arbench_res}). Inspired by ~\newcite{mccann2018natural} and~\newcite{wang2018glue} who score NLP systems based on their performance on multiple datasets, we introduce an \textbf{\textit{ARLUE score}}. The ARLUE score is simply the macro-average of the different scores across all task clusters, weighting each task equally. Following~\newcite{wang2018glue}, for tasks with multiple metrics (e.g., accuracy and F\textsubscript{1}), we use an unweighted average of the metrics as the score for the task when computing the overall macro-average. As Table~\ref{tab:Arbench_res} shows, \textit{\textbf{our MARBERT-v2 model achieves the highest ARLUE score ($77.40$)}}, followed by XLM-R\textsubscript{L} ($76.55$) and~\ourmodel~($76.07$). We also note that in spite of its superiority on social data,~\newmodel~ ranks top $4$. This is due to~\newmodel~ suffering on the QA tasks (due to its short input sequence length), and to a lesser extent on NER and TC.

\input{tables-App/tab_ArBench_res}
\input{tables/tab_ArBench_res}

%% file: tables-App/tab_ArBench_res.tex
\begin{table*}[t]
\centering
\renewcommand{\arraystretch}{1.2}
\resizebox{\textwidth}{!}{%
\begin{tabular}{lcccccccccccccc}
\toprule

\textbf{Dataset} & \textbf{mBERT} & \textbf{XLM-R\textsubscript{B}} & \textbf{XLM-R\textsubscript{L}} & \textbf{AraBERT} & \textbf{ARBERT} & \textbf{MARBERT} & \textbf{MARBERT (v2)} \\

\toprule

ARLUE\textsubscript{Senti}$^\star$   & $79.02$ / $79.50$                      & $92.17$ / $93.00$                        & $\bf93.18$ / $\bf94.00$                        & $78.26$ / $78.50$                      & $87.96$ / $88.50$                      & $\bf93.30$ / $\bf94.00$                        & $92.82$ / $93.50$                      \\
ARLUE\textsubscript{SM}\textsuperscript{$\dagger$}       & $66.84$ / $61.76$                     & $69.18$ / $64.12$                     & $68.79 $  / $64.20 $                     & $67.63$ / $62.11$                     & $69.12$ / $64.23$                     & $\bf71.64 $ / $\bf68.38$                     & $70.43$ / $66.26$                     \\
ARLUE\textsubscript{Topic}   & $91.10$ / $91.67$                     & $91.57$ / $92.24$                     & $\bf92.66$ / $\bf93.53$                     & $92.42$ / $93.17$                     & $91.06$  / $92.23$                     & $90.48$ / $92.01$                     & $91.52$ / $92.50$                      \\
ARLUE\textsubscript{Dia-B}  & $87.83$ / $87.50$                      & $88.20$ / $87.93$                     & $88.92$  / $88.57$                     & $89.30$  / $89.06$                     & $89.53$ / $89.23$                     & $89.80$ / $89.50$                      & $\bf90.05$  / $\bf89.72 $                    \\
ARLUE\textsubscript{Dia-R}  & $86.45$ / $85.89 $                    & $86.00$  / $85.46 $                    &$86.97$ / $86.54$                     & $87.30$ / $86.92$                     & $88.85$ / $88.49$                     & $\bf90.94$ / $\bf90.65$                     & $90.04$ / $89.67$                     \\
ARLUE\textsubscript{Dia-C} & $41.08$ / $32.03$                     & $40.59$ / $31.75$                     & $39.73$ / $31.51$                     & $37.90$ / $30.41$                     & $42.51$ / $34.26$                     & $43.54$ / $34.25$                     & $\bf45.37$ / $\bf35.94$                     \\
ARLUE\textsubscript{NER}     & $96.81$ / $76.91$                     & $97.74$ / $84.09$                     & $\bf97.97$ / $\bf85.56$                     &$97.79$ / $83.67$                     & $97.46$ / $81.21$                     & $96.89$ / $76.58$                     & $97.18$ / $79.34$                     \\
			
ARLUE\textsubscript{QA}$^\ddagger$     & $32.30$ / $51.14$                     & $32.30$ / $52.43$                     & $35.18$ / $\bf58.08$                     &$31.72$ / $51.87$                     & $34.04$ / $54.34$                     & $27.27$ / $43.67$                     & $\bf37.14$ / $57.93$                     \\
	
\hline
Average  & $72.68$ / $70.80$ & $74.72$ / $73.88$ & $75.43$ / $75.79$ & $75.75$ / $71.96$ & $75.07$ / $74.06$ & $75.48$  / $73.63$ & $\bf76.82$ / $\bf75.61$\\
\toprule
			
\textbf{ARLUE\textsubscript{Score}}  & $71.74$ & $74.30$ & $75.34$ & $72.38$ & $74.56$ & $74.56$ & $\bf76.21$
\\
\toprule
\end{tabular}%
}


\caption{\small{Performance of our models on the \textbf{DEV} splits of ARLUE. $^\star$ Metric for ARLUE\textsubscript{Senti} is F\textsubscript{1}\textsuperscript{PN}. $^\dagger$ ARLUE\textsubscript{SM} results is the average score across the social meaning tasks described in Table~\ref{tab:appedix_SM_res}}. $^\ddagger$~Metric for ARLUE\textsubscript{QA} is Exact Match (EM) / F$_1$.}
\label{tab:appendix_Arbench_res}
\end{table*}

%% file: tables/tab_ArBench_res.tex
\begin{table*}[]
\centering
\renewcommand{\arraystretch}{1.2}
\resizebox{\textwidth}{!}{%
\begin{tabular}{lcccccccccccccc}
\toprule
                         
\textbf{Dataset}&\textbf{mBERT} & \textbf{XLM-R\textsubscript{B}} & \textbf{XLM-R\textsubscript{L}} & \textbf{AraBERT} & \textbf{ARBERT} & \textbf{MARBERT} & \textbf{MARBERT (v2)} \\
\toprule
ARLUE\textsubscript{Senti}$^\star$           & $79.02$ / $79.50$       & $92.17$ / $93.00$       & $93.18$ / $\bf 94.00$       & $78.26$ / $78.50$        & $87.96$ / $88.50$       & $\bf 93.30$  / $\bf 94.00$        & $\bf 93.30$ / $\bf 94.00$          \\
ARLUE\textsubscript{SM}\textsuperscript{$\dagger$}              & $77.76$ / $69.88$       & $79.81$ / $71.19$       & $80.01$ / $73.00$       & $78.84$ / $72.03$        & $80.39$ / $74.22$       & $\bf82.35$ / $\bf77.13$       & $76.34$ / $\bf77.13$           \\
ARLUE\textsubscript{Topic}           & $90.88$ / $92.12$       & $90.90$ / $91.81$       & $\bf 92.24$ / $\bf 93.40$       & $92.15$ / $92.97$        & $90.81$ / $92.65$       & $89.67$ / $90.97$        & $90.07$ / $91.54$          \\
ARLUE\textsubscript{Dia-B}          & $85.52$ / $84.88$       & $86.54$ / $85.98$       & $87.82$ / $87.17$       & $87.74$ / $87.21$        & $88.31$ / $87.74$       & $\bf88.72$ / $\bf88.19$        & $ 88.47$ /  $87.87$          \\
ARLUE\textsubscript{Dia-R}          & $86.45$ / $85.89$       & $86.00$ / $85.46$       & $86.97$ / $86.54$       & $87.30$ / $86.92$        & $88.85$ / $88.49$       & $\bf 90.94$ / $\bf 90.65$        &$90.04$ / $89.67$            \\
ARLUE\textsubscript{Dia-C}         & $42.80$ / $35.23$       & $42.67$ / $35.40$       & $41.94$ / $34.98$       & $39.71$ / $33.56$        & $44.44$ / $36.87$       & $45.89$ / $37.69$        & $\bf 47.49$ / $\bf38.53$          \\
ARLUE\textsubscript{NER}             & $95.90$ / $69.06$       & $96.02$ / $73.27$       & $96.13$ / $\bf 74.94$       & $96.76$ / $76.19$        & $97.00$ / $76.83$       & $96.38$ / $71.93$        & $\bf 96.75$ / $74.70$          \\
ARLUE\textsubscript{QA}$^\ddagger$              & $34.34$ / $55.74$       & $34.62$ / $56.67$       & $39.37$ / $\bf 63.12$       & $36.31$ / $58.10$        & $36.29$ / $57.81$       & $29.13$ / $48.83$        & $\bf 40.47$ / $62.09$          \\
\hline
Average  & $74.08$ / $71.54$       & $76.09$ / $74.10$       &$77.21$ / $75.89$       & $74.63$ / $73.19$        & $76.76$ / $75.39$       & $77.05$ / $74.92$        & $\bf77.87$  / $\bf 76.94$    \\
	\toprule		
\textbf{ARLUE\textsubscript{Score}}  & $72.81$       & $75.09$       &$76.55$       & $73.91$        & $76.07$       & $75.99$        & $\bf77.40$    \\
\toprule
\end{tabular}%
}
\caption{\small{Performance of our models on the \textbf{TEST} splits of ARLUE (Acc / F$_1$). $^\star$ Metric for ARLUE\textsubscript{Senti} is Acc/ F\textsubscript{1}\textsuperscript{PN}. $^\dagger$ ARLUE\textsubscript{SM} results is the average score across the social meaning tasks described in Table~\ref{tab:SM_res}.} $^\ddagger$~Metric for ARLUE\textsubscript{QA} is Exact Match (EM) / F$_1$. }
\label{tab:Arbench_res}
\end{table*}

%% file: lit.tex
\section{Related Work}\label{sec:lit}

\note{\hl{Nice part about word embeddings and contextual representation commented below}. 

Distributed representations of words, as in Word2Vec~\cite{mikolov2013distributed}, GloVe~\cite{pennington2014glove} and FastText~\cite{mikolov2017advances}, have brought significant improvements to NLP. Contextualized word embeddings such as ELMo~\cite{peters2018deep} and Flair~\cite{akbik2018contextual} have made it possible to provide more context-sensitive (hence more accurate) representations of words, and a growing list of embeddings~\cite{akbik2019flair} followed.}

\noindent\textbf{English and Multilingual LMs.} Pre-trained LMs exploiting a self-supervised objective with masking such as BERT~\cite{devlin2019bert} and RoBERTa~\cite{liu2019incorporating} have revolutionized NLP. Multilingual versions of these models such as mBERT and XLM-RoBERTa~\cite{conneau-etal-2020-unsupervised} were also pre-trained. Other models with different objectives and/or architectures such as ALBERT~\cite{lan2019albert}, T5~\cite{raffel2019exploring} and its multilingual version, mT5~\cite{xue2020mt5}, and GPT3~\cite{brown2020language} were also introduced. More information about BERT-inspired LMs can be found in~\newcite{rogers2021primer}.


\noindent\textbf{Non-English LMs.} Several models dedicated to individual languages other than English have been developed. These include AraBERT~\cite{antoun2020arabert} and ArabicBERT~\cite{safaya2020} for Arabic, Bertje for Dutch~\cite{de2019bertje}, CamemBERT~\cite{martin-etal-2020-camembert} and FlauBERT~\cite{le2020flaubert} for French, PhoBERT for Vietnamese~\cite{nguyen2020phobert}, and the models presented by~\newcite{virtanen2019multilingual} for Finnish,~\newcite{dadas2020pre} for Polish, and~\newcite{malmsten2020playing} for Swedish.~\newcite{pyysalo2020wikibert} also create monolingual LMs for 42 languages exploiting Wikipedia data. Our models contributed to this growing work of dedicated LMs, and has the advantage of covering a wide range of dialects. Our ~\newmodel~and MARBERT-v2 models are also trained with a massive scale social media dataset, endowing them with a remarkable ability for real-world downstream tasks.

\noindent\textbf{NLP Benchmarks.} 
In recent years, several NLP benchmarks were designed for comparative evaluation of pre-trained LMs. For English,~\newcite{mccann2018natural} introduced NLP Decathlon  (DecaNLP) which combines $10$ common NLP datasets/tasks.  ~\newcite{wang2018glue} proposed GLUE, a popular benchmark for evaluating nine NLP tasks. \newcite{wang2019superglue} also presented SuperGLUE, a more challenging benchmark than GLUE covering seven tasks. In the cross-lingual setting, \newcite{pmlr-v119-hu20b} provide a Cross-lingual TRansfer Evaluation of Multilingual Encoders (XTREME) benchmark for the evaluation of cross-lingual transfer learning covering nine tasks for $40$ languages ($12$ language families). \textit{\textbf{ARLUE complements these benchmarking efforts, and is focused on Arabic and its dialects. ARLUE is also diverse (involves $42$ datasets) and challenging (our best ARLUE score is at $77.40$).}}

\note{\hl{We could add the commented info. below about lang models in the appendices.}
}
 
 

 

%% file: conc.tex
\section{Conclusion}\label{sec:conclusion} 

We presented our efforts to develop two powerful Transformer-based language models for Arabic. Our models are trained on large-to-massive datasets that cover different domains and text genres, including social media. By pre-training \newmodel~and MARBERT-v2 on dialectal Arabic, we aim at enabling downstream NLP technologies that serve wider and more diverse communities. Our best models perform better than (or on par with) XLM-R\textsubscript{Large} ($\sim 3.4 \times$ larger than our models), and hence are more energy efficient at inference time. Our models are also significantly better than AraBERT, the currently best-performing Arabic pre-trained LM. We also introduced AraLU, a large and diverse benchmark for Arabic NLU composed of $42$ datasets thematically organized into six main task clusters. ARLUE fills a critical gap in Arabic and multilingual NLP, and promises to help propel innovation and facilitate meaningful comparisons in the field. Our models are publicly available. We also plan to publicly release our ARLUE benchmark. In the future, we plan to explore self-training our language models as a way to improve performance following~\newcite{khalifa-etal-2021-self}. We also plan to investigate developing more energy efficient models.




%% file: acknow.tex
\section*{Acknowledgements}\label{sec:acknow}
We gratefully acknowledge support from the Natural Sciences and Engineering Research Council of Canada, the Social Sciences and Humanities Research Council of Canada, Canadian Foundation for Innovation, Compute Canada and UBC ARC-Sockeye (\url{https://doi.org/10.14288/SOCKEYE}). We also thank the
Google TFRC program for providing us with free TPU access.\\

%% file: ethics.tex
\vspace{-7mm}
\section*{Ethical Considerations}\label{sec:ethics} 
Although our language models are pre-trained using datasets that were public at the time of collection, parts of these datasets might become private or get removed (e.g., tweets that are deleted by users). For this reason, we will not release or re-distribute any of the pre-training datasets. Data coverage is another important consideration: Our datasets have wide coverage, and one of our contributions is offering models that can serve more diverse communities in better ways than existing models. However, our models may still carry biases that we have not tested for and hence we recommend they be used with caution. Finally, our models deliver better performance than larger-sized models and as such are more energy conserving. However, smaller models that can achieve simply `good enough' results should also be desirable. This is part of our own future research, and the community at large is invited to develop novel methods that are more environment friendly. 

%% file: appendix.tex
\clearpage
\pagenumbering{arabic}
\twocolumn[{%
 \centering
 
}]
\appendixpage
\addappheadtotoc
\numberwithin{figure}{section}
\numberwithin{table}{section}






\section{Sentiment Analysis}\label{sec_app:SA}

\subsection{SA Datasets}
\label{subsection:SA_Datasets}

\input{datasets/SA_datasets}
\input{tables/SA_distrub}
\subsection{SA Baselines}
\label{subsection:SA_Baselines}
\input{baselines/SA_baselines}
\subsection{ SA Evaluation on DEV}
\label{subsection:SA_Evaluation} 
Table~\ref{tab:appedix_senti_res} shows results of SA on DEV for datasets where there is a development split.

\input{tables-App/tab_SA_res}

\section{Social Meaning}\label{sec_app:SM}

\subsection{SM Tasks \& Datasets}
\label{subsection:SM_Datasets}
\input{datasets/SM_datasets}

\input{tables/SM_distrub}

\subsection{SM Baselines}
\label{subsection:SM_Baselines}

\input{baselines/SM_baselines}

\subsection{ SM Evaluation on DEV}
\label{subsection:SM_Evaluation}
Table~\ref{tab:appedix_SM_res}  shows results of the social meaning tasks on development splits.
\input{tables-App/tab_SM_res}
\section{Topic Classification}\label{sec_app:TC}

\subsection{TC Datasets}
\label{subsection:TC_Datasets}

\input{datasets/TC_datasets}
\input{tables/TC_distrub} 
\subsection{TC Evaluation on DEV}
\label{subsection:TC_Evaluation}
Results of TC tasks on DEV data are in Table~\ref{tab:appedix_TC_res}.
\input{tables-App/tab_TC_res}
\section{Dialect Identification}\label{sec_app:DIA}

\subsection{DIA Datasets} \label{subsection:DIA_Datasets}

\input{datasets/DIA_datasets}

\subsection{DIA Baselines}\label{subsection:DIA_Baselines}

\input{tables/DA_distrub}

\input{baselines/DIA_baselines}

\subsection{DIA Evaluation on DEV} \label{subsection:DIA_Evaluation}
Table~\ref{tab:appedix_DIA_res} shows results of the dialect identification tasks on development splits.
\input{tables-App/tab_DIA_res}
\section{Named Entity Recognition}\label{sec_app:NER}
\subsection{NER datasets}
Table~\ref{tab:NER_distrib} and Table~\ref{tab:appedix_NER_res}  show the data splits across our NER datasets, and  the results of all our models on the development splits.

\input{tables/NER_distrub}

\subsection{NER Baselines}\label{subsec_app:NER_baseline}

\newcite{khalifa2019character} apply CNNs and BiLSTMs and report F$_1$ scores on test sets, as follows: $88.77$ (ANERcorp), $91.47$ (ACE03NW), $94.92$ (ACE03BN), $91.20$ (ACE04NW), and $65.34$ (Twitter). We use their exact data splits.

\input{tables-App/tab_NER_res}

\section{Question Answering Datasets}\label{sec_app:QA}


\input{datasets/QA_datasets}

\input{tables/QA_distrub}



%% file: datasets/SA_datasets.tex
\begin{itemize}
\item \textbf{AJGT.} The Arabic Jordanian General Tweets (AJGT) dataset~\cite{alomari2017arabic} covers MSA and Jordanian Arabic, with $900$ \textit{positive} and $900$ \textit{negative} posts.

\item \textbf{AraNET\textsubscript{Sent}}.~\newcite{mageed-2020-aranet} collect $15$ datasets in both MSA and dialects from~\newcite{mageed2012awatif} (AWATIF),~\newcite{mageed2014samar} (SAMAR),~\newcite{abdulla2013arabic,nabil2015astd,kiritchenko2016semeval,aly2013labr,salameh2015sentiment,rosenthal2017semeval,alomari2017arabic,mohammad2018semeval}, and~\newcite{baly2019arsentd}. These datasets carry both \textbf{binary} (\textit{negative} and \textit{positive}) and \textbf{three-way} (\textit{negative}, \textit{neutral}, and \textit{positive}) labels, but~\newcite{mageed-2020-aranet} map them into binary sentiment only.

\item \textbf{AraSenTi-Tweet}. This comprises $17,573$ gold labeled MSA and Saudi Arabic
tweets by~\newcite{al2017arasenti}. 

\item \textbf{ArSarcasm\textsubscript{Sent}} This sarcasm dataset is labeled with sentiment tags by~\newcite{farha2020arabic} who extract it from ASTD \cite{nabil2015astd} ($10,547$ tweets) and SemEval-2017 Task 4~\cite{rosenthal2017semeval} ($8,075$ tweets).
 
\item \textbf{ArSAS}. This Arabic Speech Act and Sentiment (ArSAS) corpus~\cite{elmadany2018arsas} consists of tweets annotated with sentiment tags.

\item \textbf{ArSenD-Lev}. The Arabic Sentiment Twitter Dataset for LEVantine dialect (ArSenD-Lev)~\cite{baly2019arsentd} has $4,000$ tweets retrieved from the Levant region.

\item \textbf{ASTD.} This is a collection of $10,006$  Egyptian tweets by~\newcite{nabil2015astd}.

\item \textbf{AWATIF}. This is an MSA dataset from newswire, Wikipedia, and web fora introduced by~\newcite{mageed2012awatif}.

\item \textbf{BBNS \&  SYTS}.  The {\bf BBN} blog posts {\bf S}entiment (BBNS) and {\bf S}yria {\bf T}weets {\bf S}entiment  (SYTS)  are introduced by \newcite{salameh2015sentiment}. 

\item \textbf{CAMel\textsubscript{Sent}}.~\newcite{obeid2020camel} merge training and development data from ArSAS~\cite{elmadany2018arsas}, ASTD~\cite{nabil2015astd}, SemEval~\cite{rosenthal2017semeval}, and ArSenTD~\cite{al2017arasenti} to create a new training dataset ($\sim	24$K) and evaluate on the independent test sets from each of these sources.

\item \textbf{HARD}. The Hotel Arabic Reviews Dataset (HARD)~\cite{elnagar2018hotel} consists of $93,700$ MSA and dialect hotel reviews.

\item \textbf{LABR}. The Large Arabic Book Review Corpus~\cite{aly2013labr} has $63,257$ book reviews from Goodreads,\footnote{\href{www.goodreads.com}{www.goodreads.com}.} each rated with a $1$-$5$ stars scale. 

\item \textbf{Twitter$_{\bf Abdullah}$.}\footnote{For ease of reference, we assign a name to this and other unnamed datasets.} This is a dataset of $2,000$ MSA and Jordanian Arabic tweets manually labeled by \newcite{abdulla2013arabic}. 

\item \textbf{Twitter$_{\bf Saad}$.} This dataset is collected using an emoji lexicon by Moatez Saad in $2019$ and is available on Kaggle.\footnote{\href{https://www.kaggle.com/mksaad/arabic-sentiment-twitter-corpus}{www.kaggle.com/mksaad/arabic-sentiment-twitter-corpus}.}

\note[mam]{Please add this info.}

\note[mam]{Where is SAMAR dataset here?}

\item \textbf{SemEval-2017}. This is the SemEval-2017 sentiment analysis in Arabic Twitter task datasetby~\newcite{rosenthal2017semeval}.

\note[mam]{Where is SYST dataset here? OK I see it is listed with BBN.}

\end{itemize}

%% file: tables/SA_distrub.tex
\begin{table}[]
\centering
\resizebox{\columnwidth}{!}{%
\begin{tabular}{llrrr}
 \toprule
\multicolumn{1}{l}{\textbf{Dataset  (classes)}} & \multicolumn{1}{l}{\textbf{Classes}}                           & \multicolumn{1}{c}{\textbf{TRAIN}} & \multicolumn{1}{c}{\textbf{DEV}} & \multicolumn{1}{c}{\textbf{TEST}} \\  \toprule
AJGT (2)                                             & \{neg, pos\}                      & $1.4$K   &-~~& $361$    \\
AraNET\textsubscript{Sent} (2)      & \{neg, pos\}                       & $100.5$K & $14.3$K & $11.8$K \\
AraSenTi-Tweet (3)                           & \{neg, neut, pos\}                                 & $11.1$K  & $1.4$K  & $1.4$K  \\
ArSar\textsubscript{Sent} (3)   & \{neg, neut, pos\}                                 & $8.4$K   &-~~& $2.1$K  \\
ArSAS (3)                                            & \{neg, neut, pos\}                                 & $24.8$K  &-~~& $3.7$K  \\
ArSenD-LEV (5)                                      & \{neg, neut, pos, neg\textsuperscript{+}, pos\textsuperscript{+}\} & $3.2$K   &-~~& $801$    \\
ASTD (3)                                             & \{neg, neut, pos\}                                 & $24.8$K  &-~~& $664$    \\
ASTD-B (2)                                            & \{neg, pos\}                      & $1.1$K   &-~~& $267$    \\
AWATIF (4)                                            & \{neg, neut, obj, pos \}                         & $2.3$K   & $288$    & $284$    \\
BBN (3)                                               & \{neg, neut, pos\}                                 & $960$     & $125$    & $116$    \\
HARD (2)                                             & \{neg, pos\}                      & $84.5   $K  &-~~& $21.1$K \\
LABR (2)                                             & \{neg, pos\}                      & $13.2$K  &-~~& $3.3$K  \\
SAMAR (5)                                             & \{mix, neg, neut, obj, pos\}                   & $2.5$K   & $310$    & $316$    \\
SemEval (3)                                          & \{neg, neut, pos\}                                 & $24.8$K  &-~~& $6.1$K  \\
SYTS (3)                                              & \{neg, neut, pos\}                                 & $960$     & $202$    & $199$    \\
Tw\textsubscript{Abdullah} (2) & \{neg, pos\}                      & $1.6$K   & $202$    & $190$    \\
Tw\textsubscript{Saad} (2)     & \{neg, pos\}                      & $47$K  & $5.8$K  & $5.8$K \\
\cdashline{1-5}
 \textbf{ARLUE\textsubscript{Senti} (3)}    & \{neg, pos,  neut\}                      & $190.9$K  & $6.5$K  & $44.2$K 
\\  \toprule
\end{tabular}%
}
\caption{\small{Sentiment analysis datasets. \textbf{neg\textsuperscript{+}}: ``very negative";  \textbf{pos\textsuperscript{+}}: ``very positive". We construct ARLUE\textsubscript{Senti} by merging the different datasets and collapsing, or removing, the less frequent classes (details in text).}}
\label{tab:SA_distrib}
\end{table}

%% file: baselines/SA_baselines.tex
For SA, we compare to the following STOA: 

\begin{itemize}

    \item \textbf{\newcite{antoun2020arabert}.} We compare to best results reported by the authors on five SA datasets: HARD, balanced ASTD (which we refer to as ASTD-B), ArSenTD-Lev, AJGT, and the unbalanced positive and negative classes for LABR. They split each dataset into 80/20 for Train/Test, respectively, and report in accuracy using the best epoch identified on test data. For a valid comparison, we follow their data splits and evaluation set up.
    
    


\item \textbf{\newcite{obeid2020camel}.} They fine-tune mBERT and AraBERT on the merged CAMel\textsubscript{sent} datasets and report in $F_1$\textsuperscript{\textit{PN}}, 
which is the macro $F_1$ score over the positive and negative classes only (while neglecting the neutral class).



\item \textbf{\newcite{mageed-2020-aranet}.}  They fine-tune mBERT on the AraNET\textsubscript{Sent} data and report results in $F_1$ score on test data.

\end{itemize}

%% file: tables-App/tab_SA_res.tex
\begin{table}[ht]
\centering
\renewcommand{\arraystretch}{1.2}
\resizebox{\columnwidth}{!}{%
\begin{tabular}{lccccccc}
\toprule
\textbf{Dataset (classes)} &\textbf{mBERT} & \textbf{XLM-R\textsubscript{B}} & \textbf{XLM-R\textsubscript{L}} & \textbf{AraBERT} & \textbf{ARBERT} & \textbf{MARBERT} \\ \toprule
AraNET\textsubscript{Sent}(2)       & $84.00$& $92.00$         & $\bf93.00$ & $86.50$& $89.00$         & $92.00$         \\
AraSenTi(3) & $93.00$& $93.50$         & $\bf95.00$& $91.50$& $92.00$         & $93.50$         \\
BBN(3)               & $68.00$& $75.00$         & $77.00$         & $70.00$& $\textbf{79.50}$ & $78.50$         \\
SYTS(3)              & $62.00$& $\textbf{80.50}$ & $66.00$         & $65.00$& $69.00$         & $72.50$         \\
Twitter\textsubscript{Saad}(2)      & $80.00$& $95.50$         & $95.50$         & $81.50$& $90.00$         & $\textbf{96.00}$ \\
SAMAR(5)             & $26.00$& $54.50$         & $61.00$         & $42.50$& $50.50$         & $\textbf{62.50}$ \\
AWATIF(4)            & $63.50$& $62.00$         & $67.50$         & $65.00$& $70.50$         & $\textbf{72.00}$ \\
Twitter\textsubscript{Abdullah}(2)  & $87.50$& $91.00$         & $95.50$         & $92.50$& $\textbf{99.00}$ & $97.00$\\ 

\toprule

\end{tabular}%
}
\caption{SA results (F\textsubscript{1}) on DEV.}
\label{tab:appedix_senti_res}
\end{table}

%% file: datasets/SM_datasets.tex
\begin{itemize}
    \item \textbf{Age and Gender}. For both age and gender, we use the \textit{Arap-Tweet} dataset~\cite{zaghouani2018arap}, which covers $17$ different countries from $11$ Arab regions. 
We follow the 80-10-10 data split of AraNet~\cite{mageed-2020-aranet}.


\item \textbf{Dangerous Speech.} We use the dangerous speech \textit{AraDang} dataset from ~\newcite{alshehri2020understanding}, which is composed of tweets manually labeled with \textit{dangerous} and \textit{safe} tags.
 
\item \textbf{Offensive Language and Hate Speech}. We use manually labeled data from the shared task of offensive speech~\cite{mubarak-etal-2020-overview}.\footnote{\href{http://edinburghnlp.inf.ed.ac.uk/workshops/OSACT4/}{http://edinburghnlp.inf.ed.ac.uk/workshops/OSACT4}.} The shared task is divided into two sub-tasks: \textbf{sub-task A}: detecting if a tweet is \textit{offensive} or \textit{not-offensive}, and \textbf{sub-task B}: detecting if a tweet is \textit{hate-speech} or \textit{not-hate-speech}. 

\item \textbf{Emotion.} We use the \textit{AraNeT\textsubscript{emo}} dataset from~\newcite{mageed-2020-aranet}, which is created by merging two datasets from~\newcite{alhuzali2018enabling}. 

\item \textbf{Irony}. We use the irony identification dataset for Arabic tweets released by IDAT@FIRE2019 shared task~\cite{idat2019}, following~\newcite{mageed-2020-aranet} data splits. 

\item \textbf{Sarcasm}. We use the \textit{ArSarcasm} dataset developed by~\newcite{farha2020arabic}. 
\end{itemize}

More details  about these datasets are in Table~\ref{tab:SM_distrib}.


%% file: tables/SM_distrub.tex

\begin{table}
\centering
\caption{Social Meaning datasets.}
\label{tab:SM_distrib}
\resizebox{0.5\textwidth}{!}{%
\begin{tabular}{lllrrr}
\toprule
\textbf{Task}  & \textbf{Dataset (classes)} & \textbf{Classes}      & \multicolumn{1}{l}{\textbf{TRAIN} } & \multicolumn{1}{l}{\textbf{DEV} }     & \multicolumn{1}{l}{\textbf{TEST} }  \\ 
\toprule
Age            &  Arap-Tweet (3)& \{ $\leq$ $24$\ yrs, $25-34$\ yrs, $\geq$ $35$ yrs   \}   & $1.3$M& $160.7$K  & $160.7$K             \\
Dangerous       & AraDang (2) & \{dangerous, not-dangerous\}           & $3.5$K& $616$     & $664$ \\

Emotion        & AraNET\textsubscript{Emo} (8) & \{ang, anticip, disg, fear, joy, sad, surp, trust\} & $190$K& $911$     & $942$ \\

Gender         & Arap-Tweet (2)& \{female, male\}      & $1.3$M& $160.7$K  & $160.7$K             \\
Hate Speech    & HS@OSACT (2)  & \{hate, not-hate\}            & $10$K & $1$K      & $2$K 
\\
Irony          & FIRE2019 (2)  & \{irony, not-irony\}         & $3.6$K& - & $404$ \\
Offensive      & OFF@OSACT (2) & \{offensive, not-offensive\}           & $10$K & $1$K      & $2$K  \\

Sarcasm        & AraSarcasm (2)& \{sarcasm, not-sarcasm\}        & $8.4$K& - & $2.1$K\\

\toprule
\end{tabular}%
}
\end{table}

%% file: baselines/SM_baselines.tex
\begin{itemize}
    \item  \textbf{Age and Gender}. We compare to AraNET~\newcite{mageed-2020-aranet} age and gender models, trained by fine-tuning mBERT. The authors report $51.42$ and $65.30$ $F_1$ on age and gender, respectively.
    
    \item \textbf{Dangerous Speech}. We compare to~\newcite{alshehri2020understanding}, who report a best of $59.60$ F$_1$ on test with an mBERT model fined-tuned on emotion data.
    
    \item \textbf{Emotion}. We compare to \newcite{mageed-2020-aranet}, who acquire $60.32$ $F_1$ on test with a fine-tuned mBERT.
    
    \item \textbf{Hate Speech}. The best results on the offensive and hate speech shared task~\cite{mubarak-etal-2020-overview} are at $95$ F$_1$ score and are reported by \newcite{husain-2020-osact4}, who employ heavy feature engineering with SVMs. 
    Since our focus is on methods exploiting language models, we compare to~\newcite{djandji-etal-2020-multi} who rank second in the shared task with a fine-tuned AraBERT ($83.41$ F$_1$ on test).
    
    \item \textbf{Irony}. We compare to~\newcite{zhang2019multi} who fine-tune mBERT on the irony task, with an auxiliary author profiling task, and report $82.4$  F$_1$ on test.
    
    \item \textbf{Offensive Language}. We compare to the best results on the offensive sub-task~\cite{mubarak-etal-2020-overview} reported by~\newcite{hassan-etal-2020-alt}. They propose an ensemble of SVMs, CNN-BiLSTM, and mBERT with majority voting and acquire $90.51$ F$_1$. 
    
    \item \textbf{Sarcasm}. We compare to ~\newcite{farha2020arabic} who train  a BiLSTM model using the AraSarcasm dataset, reporting $46.00$ F$_1$ score.
\end{itemize}

%% file: tables-App/tab_SM_res.tex

\begin{table}[]
\centering


\resizebox{\columnwidth}{!}{%
\renewcommand{\arraystretch}{1.2}
\begin{tabular}{lcccccc} 

\toprule
\textbf{Task (classes)} & \textbf{mBERT} & \textbf{\textbf{XLM-R\textsubscript{B}}} & \textbf{\textbf{XLM-R\textsubscript{L}}} & \textbf{AraBERT} & \textbf{ARBERT} & \textbf{MARBERT} \\
\toprule
Age (3)       & $56.33$ & $59.70$ & $53.63$ & $57.67$                      & $58.60$& $\textbf{62.19}$ \\
Dangerous (2) & $67.35$ & $65.09$ & $69.95$             & $67.73$                      & $68.58$         & $\textbf{75.50}$ \\
Emotion (8)   & $61.34$ & $72.09$ & $72.78$                      & $65.46$                      & $68.05$ & $\textbf{75.18}$ \\
Gender (2)    & $68.06$ & $71.10$ & $71.23$            & $67.61$                     & $69.97$          & $\textbf{72.81}$ \\
Hate (2)      & $75.91$ & $76.56$ & $78.00$                      & $72.09$                      & $75.01$          & $\textbf{82.91}$ \\
Irony (2)     & $81.08$ & $83.12$ & $81.29$                      & $79.12$                      & 84.83 & $\textbf{86.77}$ \\
Offensive (2) & $84.04$ & $85.26$ & $86.72$                      & $87.21$ & $88.77$ & $\bf91.68$\\

\toprule

\end{tabular}%
}
\caption{\footnotesize {SM results in F\textsubscript{1} on DEV.}}
\label{tab:appedix_SM_res}

\end{table}

%% file: datasets/TC_datasets.tex

\begin{itemize}

\item \textbf{Arabic News Text.} \newcite{chouigui2017ant} build the Arabic news text (ANT) dataset from transcribed Tunisian radio broadcasts. 
    
 \item \textbf{Khaleej}.~\newcite{abbas2011evaluation} created the Khaleej from Gulf Arabic websites.

\item \textbf{OSAC.}
~\newcite{saad2010osac} collect OSAC from news articles.
\end{itemize}

%% file: tables/TC_distrub.tex
\begin{table}[ht]
\centering
\resizebox{\columnwidth}{!}{%
\begin{tabular}{llrrr}
 \toprule
\multicolumn{1}{l}{\textbf{Dataset  (classes)}} & \multicolumn{1}{l}{\textbf{Classes}}                           & \multicolumn{1}{c}{\textbf{TRAIN}} & \multicolumn{1}{c}{\textbf{DEV}} & \multicolumn{1}{c}{\textbf{TEST}} \\  \toprule
ANT (5)                                    & \{C, E, I, ME, S, T\}                & $25.2$K                     & $3.2$K                    & $3.2$K        \\

Khallej (4)                                  & \{E, I, LOC, S\}                                     & $4.6$K                      & $570~$                     & $570~$                      \\
OSAC (10)               & \{E, F, H, HIST, L, R, RLG, SPS, S, STR\} & $18$K                     & $2.2$K                    & $2.2$K                     \\
\cdashline{1-5}
\textbf{ARLUE\textsubscript{Topic} (16) }                                & \{\textit{all classes}\}                                     & $47.7$K                      & $5.9$K                     & $5.9$K                      \\

\toprule
\end{tabular}%
}
\caption{TC datasets. \textbf{C}: culture, \textbf{E}: economy, \textbf{F}: family, \textbf{H}: health, \textbf{HIST}: history, \textbf{I}: international news, \textbf{L}: law, \textbf{LOC}, local news, \textbf{ME}: middle east, \textbf{R}: recipes, \textbf{RLG}: religion, \textbf{SPS}: space, \textbf{S}: sports, \textbf{STR}: stories, \textbf{T}: technology. }
\label{tab:TC_distrib}
\end{table}

%% file: tables-App/tab_TC_res.tex
\begin{table}[]

\footnotesize
\resizebox{\columnwidth}{!}{%
\renewcommand{\arraystretch}{1.4}
\begin{tabular}{lcccccc}
\toprule
\multicolumn{1}{c}{\textbf{Dataset (classes)}}  & \textbf{mBERT} & \textbf{\textbf{XLM-R\textsubscript{B}}} & \textbf{\textbf{XLM-R\textsubscript{L}}} & \textbf{AraBERT} & \textbf{ARBERT} & \textbf{MARBERT} \\
\toprule

ANTText (5)       & $85.04$ & $86.74$ & $87.41$ & $\textbf{87.98}$ & $87.06$          & $85.80$           \\
ANTTitle (5)      & $79.46$ & $80.77$ & $82.04$ & $\textbf{83.56}$ & $81.10$           & $82.36$          \\
ANTText+Title (5) & $87.24$ & $86.36$ & $88.45$ & $\textbf{88.76}$ & $87.27$          & $85.99$ \\
Khallej (4)       & $94.48$ & $95.32$ & $96.09$ & $95.65$          & $96.16$ & $\textbf{96.31}$ \\
OSAC (10)         & $97.87$ & $97.75$ & $97.61$ & $\textbf{97.94}$ & $97.56$          & $97.66$          \\
\toprule
\end{tabular}
}
\caption{\footnotesize{TC results tasks (F\textsubscript{1}) on DEV.}}
\label{tab:appedix_TC_res}
\end{table}

%% file: datasets/DIA_datasets.tex
We introduce each dataset briefly here and provide a description summary of all datasets in Table~\ref{tab:DA_distrib}.

\begin{itemize}
\item  \textbf{Arabic Online Commentary (AOC).} This is a repository of $3$M Arabic comments on online news~\cite{zaidan2014arabic}. It is labeled with MSA and three \textbf{regional} dialects (\textit{Egyptian}, \textit{Gulf}, and \textit{Levantine}).

\item  \textbf{ArSarcasm\textsubscript{Dia}}. This dataset is developed by~\newcite{farha2020arabic} for sarcasm detection but also carries \textbf{regional} dialect labels from the set \textit{\{Egyptian, Gulf, Levantine,  Maghrebi\}}.

\item  \textbf{MADAR}. Sub-task 2 of the MADAR shared task~\cite{bouamor2019madar}\footnote{\href{https://camel.abudhabi.nyu.edu/madar-shared-task-2019}{https://camel.abudhabi.nyu.edu/madar-shared-task-2019/.}} is focused on user-level dialect identification with manually-curated \textbf{country} labels (n=$21$). 

\item  \textbf{NADI-2020.} The first Nuanced Arabic Dialect Identification shared task (NADI 2020)~\cite{mageed-etal-2020-nadi}\footnote{\href{https://github.com/UBC-NLP/nadi}{https://github.com/UBC-NLP/nadi}.} targets \textbf{country} level (n=$21$)  as well as \textbf{province} level (n=$100$) dialects. 

\item  \textbf{QADI}. The QCRI Arabic Dialect Identification (QADI) dataset~\cite{abdelali2020arabic} is labeled at the \textbf{country} level (n=$18$). 

\end{itemize}

Details of the datasets are in Table~\ref{tab:DA_distrib}.

%% file: tables/DA_distrub.tex
\begin{table}[ht]
\centering
 \renewcommand{\arraystretch}{1.2}
\resizebox{\columnwidth}{!}{%
\begin{tabular}{lllrrr}
\toprule
\textbf{Task (classes)} & \textbf{Dataset}   & \textbf{Classes}                                                             & \multicolumn{1}{l}{\textbf{TRAIN}} & \multicolumn{1}{l}{\textbf{DEV}} & \multicolumn{1}{l}{\textbf{TEST}} \\
\toprule
AOC (2)                                            & Binary  & \{DA, MSA\}                                                                                                                                                                                                           & $86.5$K & $10.8$K & $10.8$K \\
AOC (3)                                            & Region  & \{Egypt, Gulf, Levnt\}                                                                                                                                                                                     & $35.7$K & $4.5$K  & $4.5$K  \\
AOC (4)                                            & Region  & \{Egypt, Gulf, Levnt, MSA\}                                                                                                                                                                                & $86.5$K                      & $10.8$K & $10.8$K \\
ArSarcasm\textsubscript{Dia} (5) & Regoin  & \{Egypt, Gulf, Levnt, Magreb, MSA\}                                                                                                                                                                                    & $8.4$K                       &-~~~~& $2.1$K                       \\


MADAR-TL (21)                                      & Country &\begin{tabular}[c]{@{}l@{}}\{\textit{Multiple countries}\textsuperscript{$\star$}\}\end{tabular} & $193.1$K                     & $26.6$K                      & $44$K                      \\


NADI (21)                                          & Country & \begin{tabular}[c]{@{}l@{}}\{\textit{Multiple countries}\textsuperscript{$\star$}\}\end{tabular} & $2.1$K                       & $5$K                       & $5$K   

  \\
  
QADI (18)                                          & Country & \begin{tabular}[c]{@{}l@{}}\{\textit{Multiple countries}\textsuperscript{$\dagger$}\}\end{tabular} & $497.8$K   & -                       & $3.5$K                  \\

  \cdashline{1-6}

\textbf{ARLUE\textsubscript{Dia-B} (2)}                                          & Binary & \{DA, MSA\} & $94.9$K   & $10.8$K                       & $12.9$K                  \\

\textbf{ARLUE\textsubscript{Dia-R} (4)}                                          & Region & \{Egypt, Gulf, Levnt, Magreb\} & $38.5$K   & $4.5$K                       & $5.3$K                 \\

\textbf{ARLUE\textsubscript{Dia-C} (21)}                                          & Country & \{\textit{Multiple countries}\textsuperscript{$\star$}\} & $711.9$K   & $31.5$K                       & $52.1$K                  \\
\toprule

\end{tabular}%
}
\caption{\small{Dialect datasets. \textsuperscript{$\star$} All Arab countries except Comoros.} \textsuperscript{$\dagger$} All Arab countries except Comoros,  Djibouti, Mauritania, and Somalia.  }
\label{tab:DA_distrib}
\end{table}

%% file: baselines/DIA_baselines.tex
\begin{itemize}
    \item \textbf{\newcite{elaraby-abdul-mageed-2018-deep}} report three levels of classification on AOC data: \textbf{(1) MSA vs. DA} ($87.23$ accuracy), \textbf{(2) regional} (i.e., \textit{Egyptian}, \textit{Gulf}, and \textit{Levantine}) ($87.81$ accuracy), and \textbf{(3) MSA, Egyptian, Gulf, and Levantine} (accuracy of $82.45$). Their best results are based on BiLSTM.

\item \textbf{\newcite{abdelali2020arabic}} fine-tune AraBERT on the QADI dataset. They report $60.6$ F$_1$. 

\item \textbf{\newcite{zhang2019no}} developed the top ranked system in MADAR sub-task $2$, with $48.76$ accuracy and $34.87$ F$_1$ at tweet level. 

\item \textbf{\newcite{talafha2020multi}} developed NADI sub-task 1 (\textbf{country level}) winning system, an ensemble of fine-tuned AraBERT ($26.78$ F$_1$). 

\item \textbf{\newcite{mekki:2020:weighted}} developed NADI sub-task 2 (\textbf{province level}) winning system using a combination of word and character n-grams to fine-tune AraBERT ($6.08$ F$_1$). 

\item \textbf{AraBERT.} For ArSarcasm\textsubscript{Dia}, where no dialect id system was previously developed, we consider a fine-tuned AraBERT a baseline.

\end{itemize}

%% file: tables-App/tab_DIA_res.tex
\begin{table}[]
\centering
 \renewcommand{\arraystretch}{1.2}
\resizebox{\columnwidth}{!}{%
\begin{tabular}{llccccccc}
\toprule
\textbf{Dataset (classes)} & \textbf{Task} &\textbf{mBERT} & \textbf{XLM-R\textsubscript{B}} & \textbf{XLM-R\textsubscript{L}} & \textbf{AraBERT} & \textbf{ARBERT} & \textbf{MARBERT} \\ \toprule
MADAR(21) & Country  & $33.75$ & $34.54$ & $33.28$ & $33.47$ & $39.24$ & $\bf40.61$ \\
AOC(4)    & Regoin   &$80.07$ & $78.97$ & $79.55$ & $80.85$ & $81.96$ & $\bf83.56$ \\
AOC(3)    & Regoin   & $87.07$ & $86.80$  & $88.21$ & $88.46$ & $89.57$ & $\bf91.56$ \\
AOC(2)    & Binary   & $87.89$ & $87.63$ & $88.38$ & $88.76$ & $89.32$ & $\bf89.66$ \\
NADI(21)  & Country  & $14.49$ & $17.30$  & $18.62$ & $16.18$ & $23.73$ & $\bf26.40$  \\
NADI(100) & Province & $02.32$  & $03.91$  & $4.00$  & $03.04$  & $\bf06.05$  & $05.23$  \\
\toprule

\end{tabular}%
}
\caption{ DIA results on DEV in F\textsubscript{1}.}
\label{tab:appedix_DIA_res}
\label{tab:DIA_res}
\end{table}

%% file: tables/NER_distrub.tex
\begin{table}[t]
\centering

\renewcommand{\arraystretch}{1}
\resizebox{0.8\columnwidth}{!}{%
\begin{tabular}{lrrrr}
\hline

\textbf{Dataset} &\textbf{ Tokens}& \textbf{Train}          & \textbf{DEV}   & \textbf{Test}              \\ \hline
ANERcorp    &$150.2$K& $95.5$K & $24.8$K & $29.9$K  \\ 
ACE03BN & $15.6$K& $11.6$K & $2$K & $2$K  \\
ACE03NW & $27$K& $21.3$K & $2.7$K & $3$K  \\
ACE04BN & $70.5$K& $56.5$K & $7$K & $7$K \\ 
TW-NER    & $74.8$K & $42.9$K & $7.4$K & $24.5$K  \\ \cdashline{1-5}

\textbf{ARLUE\textsubscript{NER}} & $338.3$K& $227.7$K & $44.1$K & $66.5$K \\  

\toprule
\end{tabular}}
\caption{Distribution of the Arabic NER datasets.}
\label{tab:NER_distrib}
\end{table}

%% file: tables-App/tab_NER_res.tex
\begin{table}[]

\footnotesize
\resizebox{\columnwidth}{!}{%
\renewcommand{\arraystretch}{1.2}
\begin{tabular}{lcccccc}
\toprule
\multicolumn{1}{c}{\textbf{Dataset (classes)}}  & \textbf{mBERT} & \textbf{\textbf{XLM-R\textsubscript{B}}} & \textbf{\textbf{XLM-R\textsubscript{L}}} & \textbf{AraBERT} & \textbf{ARBERT} & \textbf{MARBERT} \\
\toprule
ANERcorp        & $86.20$ & $87.24$ & $89.64$ & $\textbf{90.24}$ & $83.24$  & $80.86$          \\
ACE03NW       & $80.57$ & $88.21$ & $\textbf{90.49}$ & $89.76$          & $88.17$ & $85.02$ \\
ACE03BN       & $80.35$ & $80.36$ & $83.39$ & $81.05$ & $\bf90.91$          & $79.05$           \\
ACE04NW      & $87.21$ & $90.08$ & $\bf91.94$ & $89.70$ & $89.33$           & $86.80$          \\
TW-NER & $52.60$ & $73.61$ & $\textbf{77.70}$ & $\textbf{73.61}$ & $70.78$          & $67.39$ \\

\toprule
\end{tabular}
}
\caption{\footnotesize{ NER results (F\textsubscript{1}) on DEV.}}
\label{tab:appedix_NER_res}
\end{table}

%% file: datasets/QA_datasets.tex

\begin{itemize}

\item \textbf{ARCD.}~\newcite{mozannar2019neural} use crowdsourcing to develop the Arabic Reading Comprehension Dataset. We use the same ARCD data splits used by ~\newcite{antoun2020arabert}.

\item \textbf{MLQA.}  This MultiLingual Question Answering benchmark is proposed by ~\newcite{lewis2019mlqa}. It consists of over $5$K extractive question-answer  instances in SQuAD format in seven languages, including Arabic. 

\item \textbf{XQuAD.} This Cross-lingual Question Answering Dataset~\newcite{artetxe2020cross} consists of $1,190$ question-answer pairs and $240$ paragraphs from SQuAD v1.1~\cite{rajpurkar2016squad} translated into ten languages (including Arabic) by professional translators. 

\item \textbf{TyDi QA.} The TyDi QA dataset~\newcite{artetxe2020cross} is manually curated and covers $11$ languages (including Arabic).  We focus on the ``Gold" passage task only.
\end{itemize}

%% file: tables/QA_distrub.tex
\begin{table}[h!]
\centering
\tiny 
\renewcommand{\arraystretch}{1.1}
\begin{tabular}{lccc}
\hline

\textbf{Dataset} & \textbf{TRAIN}   & \textbf{DEV} & \textbf{TEST}             \\ \hline

AR-XTREME & $86.7$K (MT) & -  &  - \\
ARCD & -  &  - &  $1.4$K (H) \\ 
AR-MLQA  & -  & $517$ (HT) &  $5.3$K (HT) \\ 
AR-XQuAD  & -  &  - &  $1.2$K (HT)\\ 
AR-TyDi-QA  & $14.8$K (H) & -  &  ~$~921$ (H)  \\


 \cdashline{1-4}

\textbf{ARLUE\textsubscript{QA}} & $101.6$K & $517$  & $11.6$K  \\  
\toprule
\end{tabular}
\caption{Multilingual \& Arabic QA  datasets. \textbf{H}: Human Created. \textbf{HT}: Human Translated. \textbf{MT}: Machine Translated.  }
\label{tab:QA_distrib}
\end{table}